\documentclass[sigconf,nonacm]{acmart}
\AtBeginDocument{%
  }

\setcopyright{none}
\acmSubmissionID{393}

\usepackage{array}
\usepackage[caption=false,font=normalsize,labelfont=sf,textfont=sf]{subfig}
\usepackage{textcomp}
\usepackage{verbatim}
\usepackage{multirow}
\usepackage{pifont}
\usepackage{amsmath,amsbsy}
\usepackage{bm}
\usepackage{mathtools}
\usepackage{ragged2e}
\usepackage{booktabs}
\usepackage{makecell}
\usepackage[table,xcdraw,dvipsnames]{xcolor}
\usepackage{threeparttable}
\usepackage{diagbox}
\usepackage{marvosym}
\usepackage[ruled,vlined]{algorithm2e}

\definecolor{myPink}{rgb}{0.9294, 0.0078, 0.5490}
\definecolor{Gray}{gray}{0.92}
\definecolor{my_color}{HTML}{FDE9E9}

\definecolor{my_color1}{HTML}{FFEACE}
\definecolor{my_color2}{HTML}{FBEAFF}
\definecolor{my_color3}{HTML}{FFC1B5}

\begin{document}

\title{ClusIR: Towards Cluster-Guided All-in-One Image Restoration}

\author{Shengkai Hu}
\authornote{Both authors contributed equally to this research.}
\affiliation{%
  \institution{Zhongnan University of Economics and Law}
  \city{Wuhan}
  \country{China}}
\email{shengkaihu@stu.zuel.edu.cn}

\author{Jiaqi Ma}
\authornotemark[1]
\affiliation{%
  \institution{Mohamed bin Zayed University of Artificial Intelligence}
  \city{Abu Dhabi}
  \country{United Arab Emirates}}
\email{jiaqi.ma@mbzuai.ac.ae}

\author{Xu Zhang}
\affiliation{%
  \institution{Wuhan University}
  \city{Wuhan}
  \country{China}}
\email{zhangxu0802@whu.edu.cn}

\author{Yongcheng Jing}
\affiliation{%
  \institution{Nanyang Technological University}
  \city{Singapore}
  \country{Singapore}}
\email{yongcheng.jing@ntu.edu.sg}

\author{Lefei Zhang}
\affiliation{%
  \institution{Wuhan University}
  \city{Wuhan}
  \country{China}}
\email{zhanglefei@whu.edu.cn}

\author{Jun Wan}
\authornote{Corresponding author.}
\affiliation{%
  \institution{Zhongnan University of Economics and Law}
  \city{Wuhan}
  \country{China}}
\email{junwan2014@whu.edu.cn}

\renewcommand{\shortauthors}{Hu et al.}


% \renewcommand{\shortauthors}{Anonymous Authors}

\begin{abstract}
 All-in-One Image Restoration (AiOIR) aims to recover high-quality images from diverse degradations within a unified framework. However, existing methods often fail to explicitly model degradation types and struggle to adapt their restoration behavior to complex or mixed degradations. To address these issues, we propose ClusIR, a \textbf{Clus}ter-Guided \textbf{I}mage \textbf{R}estoration framework that explicitly models degradation semantics through learnable clustering and propagates cluster-aware cues across spatial and frequency domains for adaptive restoration. Specifically, ClusIR comprises two key components: a Probabilistic Cluster-Guided Routing Mechanism (PCGRM) and a Degradation-Aware Frequency Modulation Module (DAFMM). The proposed PCGRM disentangles degradation recognition from expert activation, enabling discriminative degradation perception and stable expert routing. Meanwhile, DAFMM leverages the cluster-guided priors to perform adaptive frequency decomposition and targeted modulation, collaboratively refining structural and textural representations for higher restoration fidelity. The cluster-guided synergy seamlessly bridges semantic cues with frequency-domain modulation, empowering ClusIR to attain remarkable restoration results across a wide range of degradations. Extensive experiments on diverse benchmarks validate that ClusIR reaches competitive performance under several scenarios.
\end{abstract}




\maketitle
\begin{figure}[!t]
  \centering
  \includegraphics[width=\linewidth]{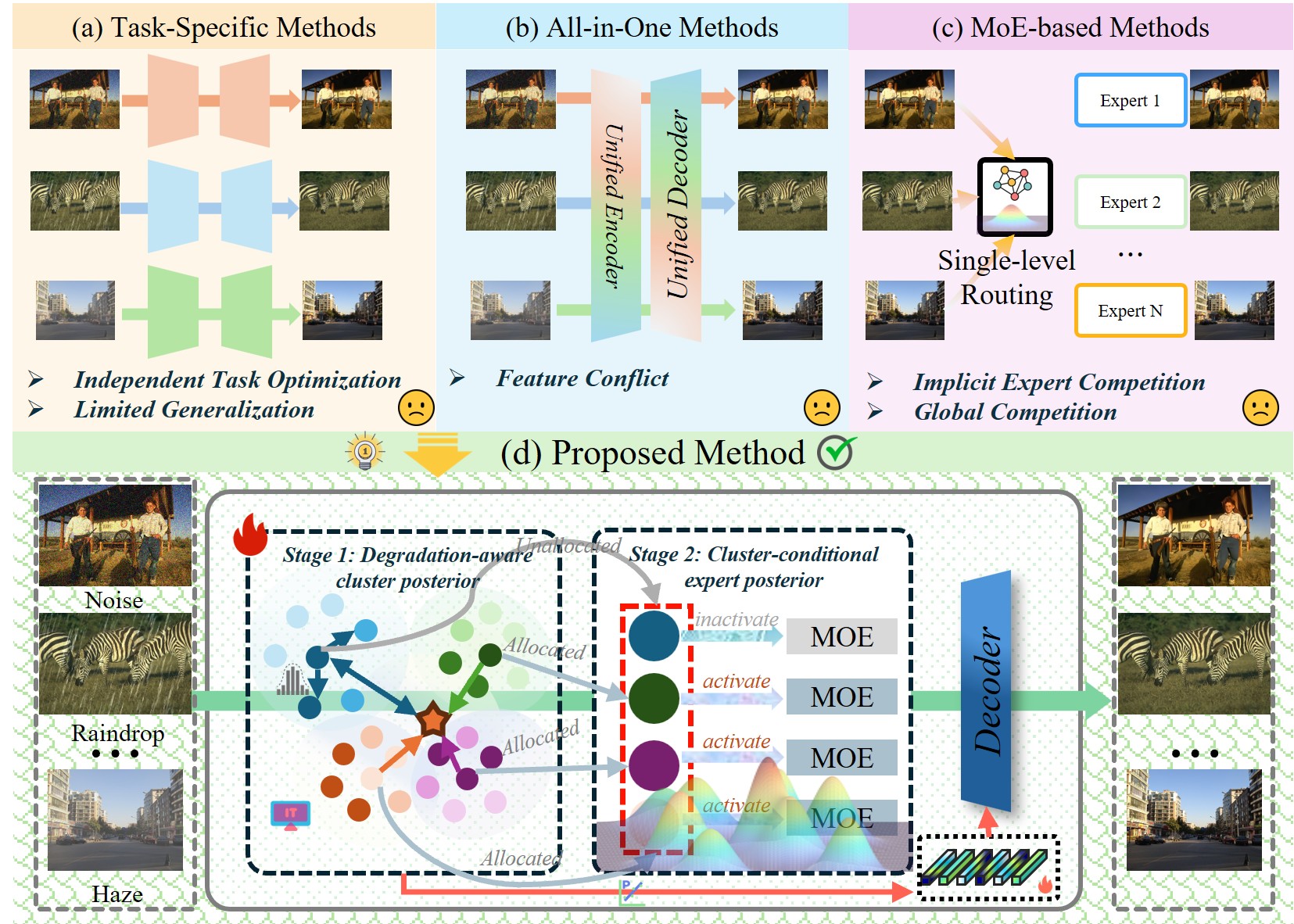}
   \vspace{-1.3em}
  \caption{
  Comparison of common image restoration paradigms.
  (a) Task-specific methods train separate networks for each degradation.
  (b) All-in-one methods unify multiple degradations within a single encoder–decoder.
  (c) MoE-based methods introduce expert specialization via dynamic routing.
  (d) Our proposed ClusIR enables explicit degradation discrimination through a Probabilistic Cluster-Guided Routing Mechanism, resulting in improved robustness to complex and mixed degradations.
  }
  \label{fig:overview}
  % \vspace{-1em}
\end{figure}

\section{Introduction}
\label{sec:intro}

Image restoration\cite{zhang2026ecrformer,ma2022elmformer,jiang2025rep,zhang2025multi,ma2025evoir} aims to recover high-quality images from degraded observations such as noise, rain, haze, blur, and low light. As a fundamental task in low-level vision, it is vital for improving the performance of downstream applications, including detection \cite{wan2023precise,hu2025,wang2025hypersigma}, tracking \cite{huang2025single}, and recognition \cite{liu2025facial}. Traditional approaches are typically task-specific, with each network tailored to a particular degradation type (Fig.~\ref{fig:overview}(a)), such as denoising \cite{DnCNN,FFDNet,ADFNet,MIRNet_v2,Restormer,NAFNet}, dehazing \cite{DehazeNet,FDGAN,DehazeFormer,FSNet}, or low-light enhancement \cite{URetinex,Retinexformer,MIRNet,DiffIR,xiao2025occlusion}. However, task-specific methods exhibit limited generalization capability, since models optimized for specific degradations often fail to adapt to unseen or complex degradation scenarios.

To address the limitations of task-specific models, recent works have explored All-in-One Image Restoration (AiOIR)\cite{zhang2023all,yin2024flexir,ProRes,PromptIR}, aiming to handle diverse degradations within a unified framework (Fig.~\ref{fig:overview}(b)). Early methods like AirNet~\cite{AirNet} introduced degradation encoders for feature extraction, while ProRes~\cite{ProRes} and PromptIR~\cite{PromptIR} employed visual prompts for adaptive guidance. More recent approaches~\cite{clip2,lin2023multi,li2025hybrid,tang2025baryir} leverage large-scale vision models (e.g., CLIP~\cite{clip1,wang2026cpl}, DINO~\cite{clip3}) to improve generalization. Despite their versatility, these unified frameworks often depend on shared representations, limiting adaptability to specific degradations and robustness under complex or hybrid scenarios.

Recently, Mixture-of-Experts (MoE) frameworks~\cite{zhang2024efficient,wang2025m2restore,zhang2024tale,yang2024language}      (Fig.~\ref{fig:overview}(c)) have enhanced AiOIR generalization by introducing expert specialization over shared representations. MEASNet~\cite{measnet} exploits pixel- and frequency-level cues to guide expert selection, while MoCE-IR~\cite{DBLP:conf/cvpr/ZamfirWMTP0T25} employs complexity-aware experts for dynamic resource allocation. MoFME~\cite{zhang2024efficient} introduces an uncertainty-aware router to achieve scalable specialization, and UniRestorer~\cite{lin2024unirestorer} adopts multi-granularity degradation representations for unified restoration. However, they rely on implicit expert competition without explicit degradation separation, leading to overlapping activations and ambiguous routing.

Motivated by the above observations, we propose ClusIR (as shown in Fig.~\ref{fig:overview}(d)), a \textbf{Clus}ter-Guided \textbf{I}mage \textbf{R}estoration framework that introduces clustering-based discrimination to decouple degraded semantics and stabilizes expert routing for adaptive image restoration. ClusIR consists of two key components: a Probabilistic Cluster-Guided Routing Mechanism (PCGRM) and a Degradation-Aware Frequency Modulation Module (DAFMM). PCGRM leverages hierarchically organized cluster prototypes to achieve discriminative and collaborative expert activation, thereby enabling explicit degradation separation and adaptive expert routing in the spatial domain.
Meanwhile, DAFMM employs cluster-guided priors to enhance frequency self-mining and promote effective interaction between low- and high-frequency representations, facilitating unified structural and textural restoration under complex degradations. 

Our main contributions are summarized as follows:
\begin{enumerate}
    \item We propose ClusIR, an All-in-One image restoration framework which incorporates explicit degradation information to bridge spatial and frequency domains. It achieves state-of-the-art performance across multiple heterogeneous benchmarks, demonstrating strong generalization and robustness to diverse degradations.
    
    \item We introduce PCGRM to leverage hierarchical degradation prototypes to disentangle degradation semantics from expert activation, thereby achieving discriminative routing and improved restoration adaptability.
    
    \item We design DAFMM to leverage cluster-guided priors for adaptive frequency learning, enabling coordinated enhancement of structural and textural components for robust image restoration under complex degradations.
    
\end{enumerate}

\section{Related Work}
\label{sec:relatedwork}
\begin{figure*}[t]
  \centering
  \setlength{\abovecaptionskip}{2pt}   % 图与caption间距
  \setlength{\belowcaptionskip}{0pt}   % caption与正文间距
  \includegraphics[width=0.9\linewidth]{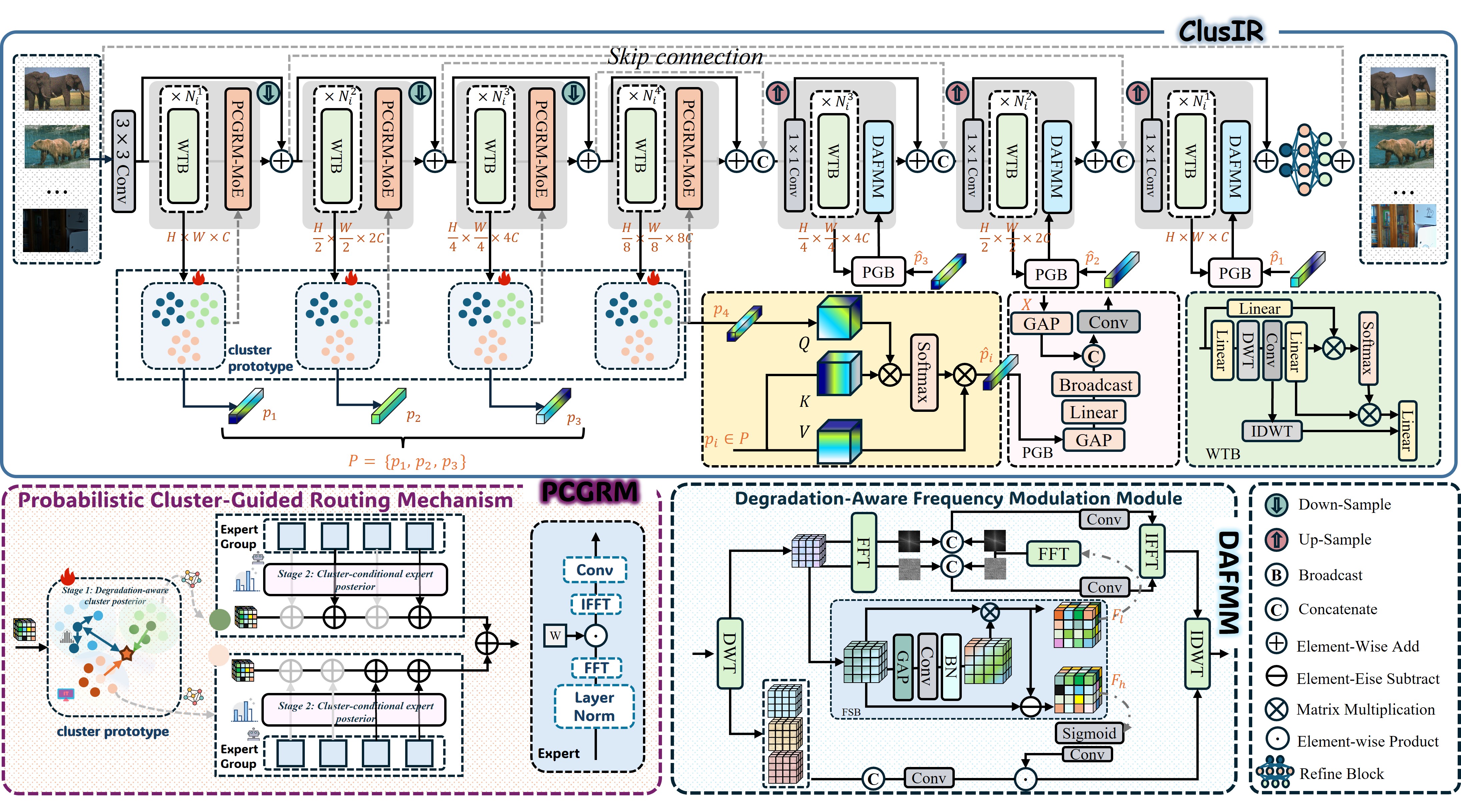}
  % \vspace{-0.6em}
  \caption{
  Overview of the proposed ClusIR framework, which integrates the Probabilistic Cluster-Guided Routing Mechanism (PCGRM) and the Degradation-Aware Frequency Modulation Module (DAFMM).
  }
  \label{fig:stru}
  \vspace{-1em}
\end{figure*}

\subsection{All-in-One image restoration}
All-in-One image restoration aims to address multiple degradation types within a unified framework. Compared to task-specific ~\cite{DehazeNet,Retinexformer,URetinex,wang2025gewdiff} and general image restoration~\cite{MIRNet_v2,MPRNet,lihe2025ada4dir}, it offers superior model efficiency and greater practicality for real-world applications. Recent studies have advanced spatial modeling and receptive-field adaptivity in unified restoration. DSwinIR~\cite{wu2025} introduces a Deformable Sliding Window Transformer with content-aware attention, while Cat-AIR~\cite{jiang2025cat} designs a content- and task-aware mechanism that balances local and global information through alternating spatial–channel attention. Beyond spatial modeling, another research line focuses on semantic adaptation through prompt learning. PromptIR~\cite{PromptIR} employs degradation-specific prompts for adaptive restoration, ProRes~\cite{ProRes} encodes multiple degradation patterns into unified visual prompts for controllable restoration, and Zhang et al.~\cite{DBLP:journals/corr/abs-2408-15994} extend this idea via a two-stage quality-aware prompting scheme.

In parallel, frequency-domain approaches aim to jointly recover structural and textural details. CSNet~\cite{csnet} integrates channel-wise Fourier transforms with multi-scale frequency modules under a frequency-aware loss, FPro~\cite{fpro} combines frequency decomposition with prompt learning for joint structural--textural enhancement, and AdaIR~\cite{DBLP:conf/iclr/0001ZKKSK25} mines degradation-specific frequency priors via bidirectional frequency--spatial modulation. However, existing frequency-based frameworks still rely on implicit feature sharing across degradations, hindering explicit degradation modeling and generalization to unseen scenarios.

\subsection{Mixture-of-Experts based image restoration}
Mixture-of-Experts (MoE) mechanisms have recently been adopted in image restoration to enhance model capacity and task adaptivity via dynamic expert routing.
Early work MEASNet~\cite{measnet} integrates frequency-domain priors into the MoE framework to jointly model spatial--frequency correlations for balanced structural--textural restoration.
Yang et al.~\cite{yang2024language} exploit textual weather descriptions to construct language-driven degradation priors for expert selection, while WM-MoE~\cite{luo2023wm} employs a weather-aware router and multi-scale experts for blind adverse-weather removal.
MoCE-IR~\cite{DBLP:conf/cvpr/ZamfirWMTP0T25} utilizes complexity-aware experts with adaptive computational units for resource allocation based on degradation difficulty, and MoFME~\cite{zhang2024efficient} designs feature-modulated experts with shared weights and an uncertainty-aware router for scalable specialization.
UniRestorer~\cite{lin2024unirestorer} introduces multi-granularity degradation representations and a corresponding MoE model for unified restoration, while M2Restore~\cite{wang2025m2restore} further advances this paradigm through a CLIP-guided MoE-based Mamba-CNN framework that unifies cross-modal priors with global--local modeling for adaptive, degradation-aware restoration.

While effective, existing paradigms exhibit inherent limitations.
In contrast, our ClusIR explicitly organizes degradation representations via cluster prototypes, enabling discriminative expert activation and adaptive frequency enhancement within a unified architecture.

\section{Methods}
\label{sec:Methods}
\subsection{Overall Pipeline}

Given a degraded input image $\mathbf{I}\in\mathbb{R}^{H\times W\times3}$, where $H$ and $W$ denote the spatial resolution, ClusIR first extracts shallow features using a $3\times3$ convolution. These features are subsequently fed into a four-stage encoder, where each stage comprises a Wavelet-based Transformer Block (WTB)~\cite{yao2022wave} and a Mixture-of-Experts (MoE) block driven by the proposed PCGRM. The PCGRM produces degradation-aware representations and hierarchical prompts (e.g., $p_1$, $p_2$, $p_3$, $p_4$). The multi-scale prompts ${p_{1}, p_{2}, p_{3}}$ are then integrated with $p_{4}$ via a multi-head cross-attention mechanism to generate semantic prompts ${\hat{p}_{1}, \hat{p}_{2}, \hat{p}_{3}}$. Each semantic prompt is refined by a lightweight Prompt Generation Block (PGB)~\cite{PromptIR} and injected into the DAFMM for frequency-domain modulation, enhancing the interaction between structural (low-frequency) and textural (high-frequency) components. The decoder progressively reconstructs the restored image from the aggregated multi-scale features, ensuring consistent spatial–frequency recovery. 

\subsection{Probabilistic Cluster-Guided Routing Mechanism}
Current All-in-One MoE methods \cite{measnet, lin2024unirestorer, wang2025m2restore} predominantly adopt a single-stage routing paradigm, which directly predicts a flat expert distribution from the input features $\mathbf{x}$:
\begin{equation}
	p(E=e \mid \mathbf{x}) = \mathrm{Softmax}(\mathbf{w}^\top \mathbf{x}),
	\label{eq:flat_moe_prob}
\end{equation}
and aggregates expert outputs as:
\begin{equation}
	\mathbf{y} = \sum_{e=1}^{M} p(E=e \mid \mathbf{x}) \cdot \mathcal{E}_e(\mathbf{x}),
	\label{eq:flat_moe_output}
\end{equation}where $e$ is the expert index, $M$ is the number of experts and $\mathcal{E}_e(\mathbf{x})$ denotes the output of the $e$-th expert. $y$ denotes the restored image and $\mathbf{w}$ is the parameter of the gating mechanism. The above routing design implicitly assumes a universal gating distribution (i.e., \textbf{a unimodal distribution}) across all degradations, ignoring their heterogeneous and compositional nature. This leads to two key issues:
(1) Mixed degradations collapsed into a single distribution.
Real-world degradations (e.g., noise + blur + haze) are inherently multi-modal, yet the single-stage softmax enforces a unimodal gating, causing semantic entanglement and limiting mixed-degradation modeling.
(2) Unstable global expert competition.
Since the router must simultaneously infer degradations and select experts within the same probability space, experts compete globally, inducing gradient interference and training instability, which degrades degradation-aware representation learning and overall restoration quality.

To overcome these limitations, we propose a novel Probabilistic Cluster-Guided Routing Mechanism (PCGRM), which decomposes the routing into a \textbf{hierarchical two-stage process} that first infers a degradation-aware cluster posterior, followed by cluster-conditional expert routing, as detailed below:
\begin{equation}
	\mathbf{y}(\mathbf{x})
	=\sum\nolimits_c p(C = c \mid \mathbf{x}) \sum\nolimits_e p(E = e \mid c, \mathbf{x}) \mathcal{E}_{e}(\mathbf{x}),
\end{equation}
where $c$ denotes the index of the latent degradation type. This decomposition disentangles degradation perception and expert selection, yielding more stable and interpretable routing. Moreover, The proposed PCGRM models \textbf{a multimodal distribution}, making it possible to address image restoration under diverse and complex degradations.

\noindent{\textbf{Stage 1: Degradation-aware cluster posterior $p\left( {C = c\left| {\mathbf{x}} \right.} \right)$.}} To achieve explicit degradation discrimination and hierarchical representation modeling, we construct a hierarchy of cluster prototypes aligned with different encoder stages. Specifically, shallow encoders employ prototypes to capture diverse degradation patterns, while deeper encoders adopt prototypes to represent more compact and abstract degradation semantics.

For each encoder layer $l$, we construct a learnable prototype bank $\mathbf{P}^{(l)} \in \mathbb{R}^{L \times D}$ to capture layer-specific degradation semantics. To enhance degradation discrimination and routing stability, we impose spherical normalization and near-orthogonality constraints on the prototype bank, ensuring that each prototype lies on a unit hypersphere with minimal pairwise correlation. For the encoded multi-scale features $\mathbf{X} = \left\{ {{{\mathbf{\hat x}}_i}} \right\}_{i = 1}^B$, where $\mathbf{\hat x}_i \in \mathbb{R}^{H \times W \times D}$ and $B$ is the batch size, global average pooling (GAP) is first applied to obtain compact feature representations, followed by a linear projection and normalization. Then, the cosine similarity between each normalized feature token $\mathbf{x}_i$ and the cluster prototype $\mathbf{P}_c^{(l)}$ at layer $l$ is computed as:
\begin{equation}
	\text{sim}(\mathbf{x}, C=c)
	= \frac{\mathbf{x}^\top \mathbf{P}_c}{|\mathbf{x}|_2 |\mathbf{P}_c|_2}.
\end{equation} For brevity, we omit the indices $i$ and $l$. The degradation-aware cluster posterior is then obtained via a softmax normalization over all cluster prototypes:
\begin{equation}
	\hat \alpha_c = \frac{\exp(\text{sim}(\mathbf{x}, C=c))}
	{\sum_{i=1}^{N} \exp(\text{sim}(\mathbf{x}, C=i))},
\end{equation}where $\hat \alpha_c$ denotes the probability that the token $\mathbf{x}$ belongs to the $c$-th degradation cluster and $N$ is the number of cluster prototypes. With estimated probability, the topK method is used to choose the most probable cluster prototypes for degradation-aware routing:
\begin{equation}
	p(C=c \mid \mathbf{x})  = \alpha_c = \frac{\exp(\text{sim}(\mathbf{x}, C=c))}
	{\sum_{ i \in \{K_1\}} \exp(\text{sim}(\mathbf{x}, C=i))},
\end{equation}where $\{K_1\}$ denotes the set of selected cluster prototypes and $\alpha_c$ denotes the probability of selecting the $c$-th cluster prototype.

% \noindent{\textbf{Stage 2: Cluster-conditional expert posterior $p\left( {E = e\left| {c,{\mathbf{x}}} \right.} \right)$.}} After obtaining the degradation-aware cluster posterior from Stage 1, we activate experts within each selected cluster to enable degradation-aware routing. Assume that each cluster prototype carries a Gaussian semantic prior 
% $\mathcal{N}(\boldsymbol{\mu}_c, \mathrm{diag}(\boldsymbol{\sigma}_c^2))$, where $\mu_c\in\mathbb{R}^{D}$ and $\mathrm{diag}(\boldsymbol{\sigma}_c^2))$ represent the semantic center and feature uncertainty of the $c$-th cluster, respectively. We generate degradation prompts $\mathcal{P}$ through reparameterization, weighted by the cluster posterior $\alpha_c$:
\noindent{\textbf{Stage 2: Cluster-conditional expert posterior $p\left( {E = e\left| {c,{\mathbf{x}}} \right.} \right)$.}} 
After obtaining the degradation-aware cluster posterior from Stage 1, we activate experts within each selected cluster to enable degradation-aware routing. 
Each cluster prototype is modeled as a Gaussian semantic distribution 
$\mathcal{N}(\boldsymbol{\mu}_c, \mathrm{diag}(\boldsymbol{\sigma}_c^2))$, 
where $\boldsymbol{\mu}_c \in \mathbb{R}^{D}$ denotes the semantic center and 
$\boldsymbol{\sigma}_c$ characterizes the feature uncertainty of the $c$-th cluster. 
We generate degradation prompts $\mathcal{P}$ via reparameterization, weighted by the cluster posterior $\alpha_c$.
\vspace{-0.5em}
\begin{equation}
	\mathcal{P}
	= \sum_{c=1}^{N} \alpha_c\big(\boldsymbol{\mu}_c + \boldsymbol{\sigma}_c \odot \boldsymbol{\epsilon}_c\big), \quad  \text{s.t.} 
	\quad \boldsymbol{\epsilon}_c \sim \mathcal{N}(\mathbf{0}, \mathbf{I}).
	\label{eq:cg_prompt}
\end{equation} 

The injected Gaussian noise models intra-cluster uncertainty and continuous degradation variations, thereby mitigating prototype collapse and facilitating robust hierarchical routing under mixed and unseen degradations. Then, the prompt interacts with features through cross-attention to form a gating mechanism and the probability distribution over all experts within that cluster $c$ can be computed as:
\begin{equation}
	\mathbf{g} = \mathrm{CrossAttn}(\mathbf{x}, \mathcal{P}, \mathcal{P}),
	\label{eq:cg_context}
\end{equation}
\vspace{-1em}
\begin{equation}
	\pi_{c} = \mathrm{Softmax}(\mathbf{W}_{c}\mathbf{g} + \mathbf{b}_{c}),
	\label{eq:expert_posterior}
\end{equation}where $\mathbf{W}_{c}$ is the learnable weight and $\mathbf{b}_{c}$ serves as an interference term. Subsequently, the top $K_2$ experts with the highest activation probabilities are selected to form the active expert subset. The probability that expert $e$ is selected within cluster $c$ can be formulated as follows:
\begin{equation}
	p(E = e \mid c, \mathbf{x})
	= \frac{\exp\!\left([\mathbf{W}_{c}\mathbf{g} + \mathbf{b}_{c}]_{e}\right)}
	{\sum_{e'=1}^{K_2} \exp\!\left([\mathbf{W}_{c}\mathbf{g} + \mathbf{b}_{c}]_{e'}\right)}.
	\label{eq:cluster_expert_softmax}
\end{equation}

This two-stage probabilistic routing mechanism disentangles degradation recognition and expert activation, enabling the model to first discover potential degradation types and then specifically select experts in each cluster. By factorizing the decision space from a single flat expert distribution (i.e., a unimodal distribution) into a structured hierarchy (i.e., a multimodal distribution), the proposed PCGRM mitigates global expert interference and enables interpretable cluster-to-expert correspondence, thereby improving robustness to mixed and unseen degradations.

\subsection{Degradation-Aware Frequency Modulation Module}
\begin{figure}[t]
  \centering
  \setlength{\abovecaptionskip}{2pt}   % 图与caption间距
  \setlength{\belowcaptionskip}{0pt}   % caption与正文间距
  \includegraphics[width=\linewidth]{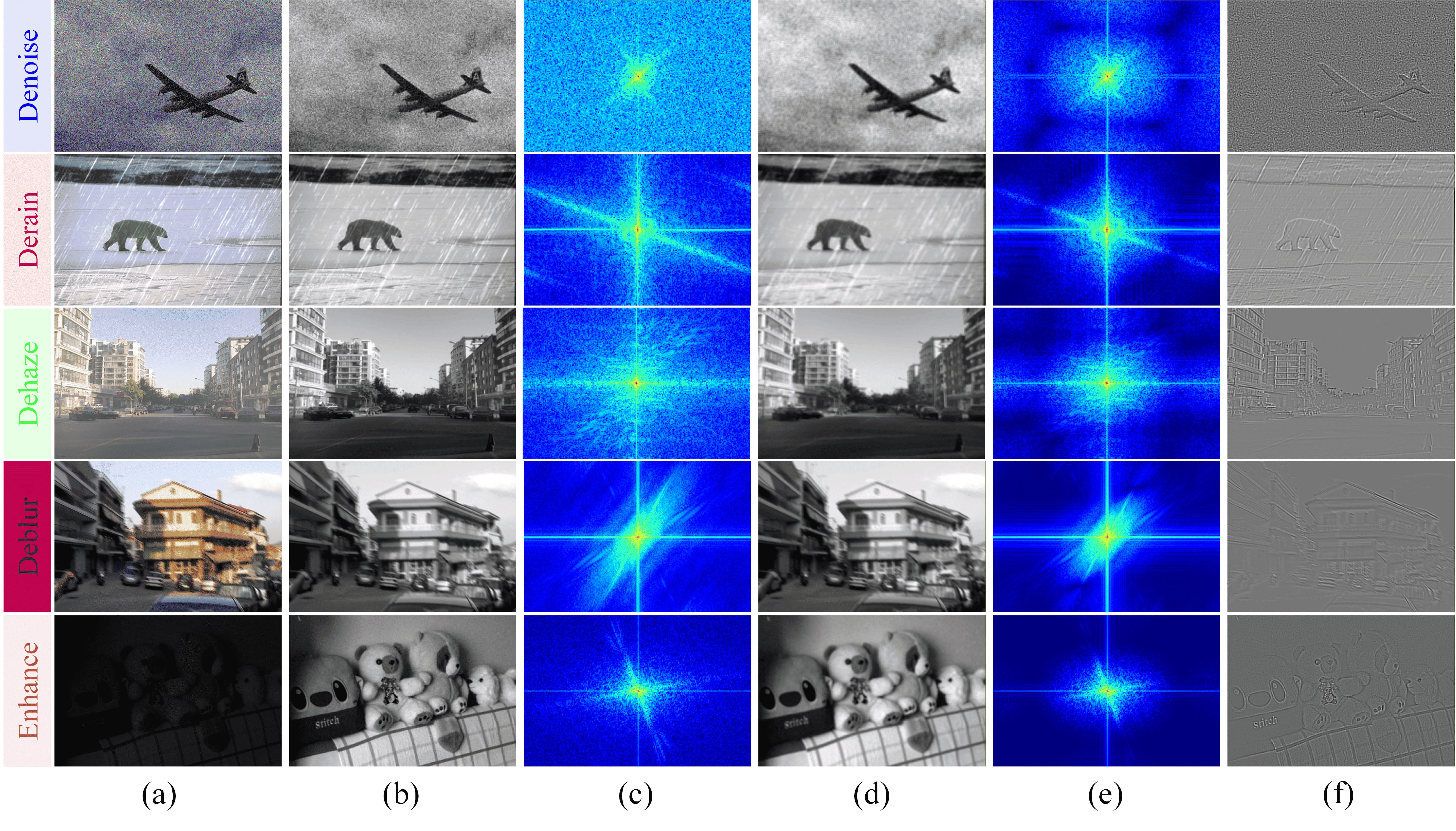}
  \caption{
  Motivation of the Frequency Self-Mining Block (FSB). (a): the input image; (b)\&(c): the low-frequency subband and frequency spectrum via DWT; (d)\&(e): the refined low-frequency response and frequency spectrum via the proposed DAFMM; (f): the high-frequency information further refined by the FSB based on (b). Our DAFMM effectively decomposes low- and high-frequency information, thereby achieving better AiOIR performance.
  }
  \vspace{-1em}
  \label{fig:fft}
\end{figure}

The proposed PCGRM explicitly decouples degradation perception and expert selection in the spatial domain, it primarily operates on global semantic representations. However, real-world degradations often exhibit frequency-dependent characteristics--for instance, high-frequency textures are more susceptible to noise, whereas low-frequency structural components are prone to blur. To bridge this gap, we extend the degradation modeling from the spatial to the frequency domain and introduce a Degradation-Aware Frequency Modulation Module (DAFMM) for fine-grained frequency enhancement. Specifically, by taking the cluster-aware semantic prompt $\mathcal{P}$ as a degradation prior, DAFMM drives a dual-branch frequency processing pathway that adaptively restores low-frequency structures and refines high-frequency textures. 

\begin{table*}[t]
\caption{Results under three AiOIR tasks (\textcolor{blue}{\bf{N}}+\textcolor{green}{\bf{H}}+\textcolor{red}{\bf{R}}) setting with SOTA methods. 
The best and second-best results are highlighted in \textbf{bold} and \underline{underlined}, respectively. 
Partial results are referenced from Perceive-IR~\cite{DBLP:journals/corr/abs-2408-15994}.}
\vspace{-0.8em}
\centering
\renewcommand\arraystretch{0.5}  % 更紧凑
\resizebox{0.95\linewidth}{!}{
\begin{tabular}{l|c|ccc|c|c|c}
\toprule[1pt]
 \multirow{2}{*}{\textbf{Method}} & \multirow{2}{*}{\textbf{Venue}} 
& \multicolumn{3}{c|}{\textbf{Denoising} (CBSD68)} 
& \multicolumn{1}{c|}{\textbf{Dehazing}} 
& \multicolumn{1}{c|}{\textbf{Deraining}} 
& \multirow{2}{*}{\textbf{Avg.}} \\
\cmidrule(lr){3-7}
& &
$\,\sigma\!=\!15$ & $\,\sigma\!=\!25$ & $\,\sigma\!=\!50$ & \multicolumn{1}{c|}{SOTS} & Rain100L & \\
\midrule
AirNet~\cite{AirNet} & CVPR'22 
& 33.92/0.932 & 31.26/0.888 & 28.00/0.797 & 27.94/0.962 & 34.90/0.967 & 31.20/0.910 \\

IDR~\cite{IDR} & CVPR'23 
& 33.89/0.931 & 31.32/0.884 & 28.04/0.798 & 29.87/0.970 & 36.03/0.971 & 31.83/0.911 \\

ProRes~\cite{ProRes} & arXiv'23 
& 32.10/0.907 & 30.18/0.863 & 27.58/0.779 & 28.38/0.938 & 33.68/0.954 & 30.38/0.888 \\

PromptIR~\cite{PromptIR} & NeurIPS'23 
& 33.98/0.933 & 31.31/0.888 & 28.06/0.799 & 30.58/0.974 & 36.37/0.972 & 32.06/0.913 \\

NDR~\cite{NDR} & TIP'24 
& 34.01/0.932 & 31.36/0.887 & 28.10/0.798 & 28.64/0.962 & 35.42/0.969 & 31.51/0.910 \\

Gridformer~\cite{Gridformer} & IJCV'24 
& 33.93/0.931 & 31.37/0.887 & 28.11/0.801 & 30.37/0.970 & 37.15/0.972 & 32.19/0.912 \\

InstructIR~\cite{InstructIR} & ECCV'24 
& \textbf{34.15}/0.933 & 31.52/0.890 & \underline{28.30}/0.804 & 30.22/0.959 & 37.98/0.978 & 32.43/0.913 \\

Up-Restorer~\cite{DBLP:conf/aaai/LiuYL025} & AAAI'25 
& 33.99/0.933 & 31.33/0.888 & 28.07/0.799 & 30.68/0.977 & 36.74/0.978 & 32.16/0.915 \\

Perceive-IR~\cite{DBLP:journals/corr/abs-2408-15994}& TIP'25 
& 34.13/0.934 & \textbf{31.53}/0.890 & \textbf{28.31}/0.804 & 30.87/0.975 & 38.29/0.980 & 32.63/0.917\\

AdaIR~\cite{DBLP:conf/iclr/0001ZKKSK25} & ICLR'25 
& 34.12/\underline{0.935} & 31.45/0.892 & 28.19/0.802 & 31.06/0.980 & 38.64/\underline{0.983} & 32.69/0.918 \\

MoCE-IR~\cite{DBLP:conf/cvpr/ZamfirWMTP0T25} & CVPR'25 
& 34.11/0.932 & 31.45/0.888 & 28.18/0.800 & 31.34/0.979 & 38.57/\textbf{0.984} & 32.73/0.917 \\

DFPIR~\cite{DBLP:conf/cvpr/TianLLLR25} & CVPR'25 
& \underline{34.14}/\underline{0.935} & \underline{31.47}/\underline{0.893} & 28.25/\underline{0.806} & \underline{31.87}/\underline{0.980} & \underline{38.65}/0.982 & \underline{32.88}/\underline{0.919} \\

\rowcolor{my_color}
\textbf{ClusIR} & Ours 
& 34.10/\textbf{0.938} & 31.45/\textbf{0.897} & 28.19/\textbf{0.814} & \textbf{32.85}/\textbf{0.983} & \textbf{38.71}/\textbf{0.984} & \textbf{33.06}/\textbf{0.923} \\
\bottomrule[1pt]
\end{tabular}}
\label{tab:three_task}
\vspace{-1.3em}
\end{table*}

\begin{table*}[!htbp]
\caption{Results under five AiOIR tasks (\textcolor{blue}{\textbf{N}}+\textcolor{green}{\textbf{H}}+\textcolor{red}{\bf{R}}+\textcolor{purple}{\textbf{B}}+\textcolor{pink}{\textbf{L}}) setting with SOTA methods.
Denoising reports only $\sigma{=}{25}$ following \cite{IDR}.}
\vspace{-1.3em}
\centering
\renewcommand\arraystretch{0.5}
\resizebox{0.95\linewidth}{!}{
\begin{tabular}{l|c|c|c|c|c|c|c}
\toprule[1pt]
 \multirow{2}{*}{\textbf{Method}} &  \multirow{2}{*}{\textbf{Venue}}
& \textbf{Denoising} & \textbf{Dehazing} & \textbf{Deraining} & \textbf{Deblurring} & \textbf{Low-light} &  \multirow{2}{*}{\textbf{Avg.}} \\
\cmidrule(lr){3-7}
& & CBSD68 ($\sigma\!=\!25$) & SOTS & Rain100L & GoPro & LOL & \\
\midrule
TAPE~\cite{TAPE} & ECCV'22
& 30.18/0.855  & 22.16/0.861  & 29.67/0.904    & 24.47/0.763   & 18.97/0.621  & 25.09/0.801 \\

TransWeather~\cite{Transweather} & CVPR'22
& 29.00/0.841  & 21.32/0.885  & 29.43/0.905    & 25.12/0.757   & 21.21/0.792  & 25.22/0.836 \\

AirNet~\cite{AirNet} & CVPR'22
& 30.91/0.882  & 21.04/0.884  & 32.98/0.951    & 24.35/0.781   & 18.18/0.735  & 25.49/0.846 \\

IDR~\cite{IDR} & CVPR'23
& \textbf{31.60}/0.887  & 25.24/0.943  & 35.63/0.965    & 27.87/0.846   & 21.34/0.826  & 28.34/0.893 \\

PromptIR~\cite{PromptIR} & NeurIPS'23
& \underline{31.47}/0.886  & 26.54/0.949  & 36.37/0.970    & 28.71/0.881   & 22.68/0.832  & 29.15/0.904 \\

Gridformer~\cite{Gridformer} & IJCV'24
& 31.45/0.885  & 26.79/0.951  & 36.61/0.971    & 29.22/0.884   & 22.59/0.831  & 29.33/0.904 \\

InstructIR~\cite{InstructIR} & ECCV'24
& 31.40/0.887  & 27.10/0.956  & 36.84/0.973    & 29.40/\underline{0.886}   & 23.00/0.836  & 29.55/0.907 \\

Perceive-IR~\cite{DBLP:journals/corr/abs-2408-15994} & TIP'25
& 31.44/0.887  & 28.19/0.964  & 37.25/0.977    & \underline{29.46}/\underline{0.886}  & 22.88/0.833  & 29.84/0.909 \\

AdaIR~\cite{DBLP:conf/iclr/0001ZKKSK25} & ICLR'25
& 31.35/\underline{0.889} & 30.53/0.978  & \underline{38.02}/\underline{0.981}  & 28.12/0.858   & 23.00/0.845  & 30.20/0.910 \\

MoCE-IR~\cite{DBLP:conf/cvpr/ZamfirWMTP0T25} & CVPR'25
& 31.34/0.887  & 30.48/0.974  & \textbf{38.04}/\textbf{0.982}    & \textbf{30.05}/\textbf{0.899}  & 23.00/\underline{0.852} & \underline{30.58}/\textbf{0.919} \\

DFPIR~\cite{DBLP:conf/cvpr/TianLLLR25} & CVPR'25
& 31.29/\underline{0.889} & \underline{31.64}/\underline{0.979} & 37.62/0.978 & 28.82/0.873 & \textbf{23.82}/0.843 & \textbf{30.64}/\underline{0.913} \\

\rowcolor{my_color}
\textbf{ClusIR} & Ours
& 31.35/\textbf{0.894}  & \textbf{31.94}/\textbf{0.982}  & 37.28/0.980    & 28.54/0.872  & \underline{23.78}/\textbf{0.865}  & \underline{30.58}/\textbf{0.919} \\

\bottomrule[1pt]
\end{tabular}}
\label{tab:five_task}
% \vspace{-1em}
\end{table*}

The DAFMM first decomposes the degradation-aware features into low-frequency 
components (i.e., $\hat{F}_{LL}$) and high-frequency components 
(i.e., $\hat{F}_{HL}$, $\hat{F}_{LH}$, and $\hat{F}_{HH}$) via a discrete 
wavelet transform (DWT), as shown in Fig.~\ref{fig:stru}. We visualize the low-frequency subband ${\hat{F}_{LL}}$ in Fig.~\ref{fig:fft}, and find that ${\hat{F}_{LL}}$ still retains prominent edge structures (Fig.~\ref{fig:fft}(b)) and residual mid-frequency components (Fig.~\ref{fig:fft}(c)). The results reveal that the DWT cannot fully decouple structural and textural information, potentially leading to suboptimal performance in image restoration. To address this limitation, the proposed DAFMM incorporates a learnable Frequency Self-Mining Block (FSB) that adaptively disentangles global low-frequency structures from localized high-frequency details. Each frequency component is then refined through tailored modulation strategies, thereby achieving more effective image restoration.

\noindent\textbf{Frequency Self-Mining Block.}  
Given ${\hat{F}_{LL}}$, a learnable low-pass filter is constructed by leveraging global average pooling (GAP) and generating softmax-normalized dynamic weights. Then, the ${\hat{F}_{LL}}$ is unfolded into local patches and adaptively filtered to obtain the low-frequency component $\mathcal{F}_l$. The high-frequency component $\mathcal{F}_h$ is obtained in a residual manner. The whole process can be formulated as:  
\begin{equation}
	W = \mathrm{Softmax}\big( \mathrm{BN}( \mathrm{Conv}_{1\times1}(\mathrm{GAP}(\hat{F}_{LL})) ) \big), \\
\end{equation}
\begin{equation}
	X = \operatorname{Unfold}(\hat{F}_{LL}), \;
	\mathcal{F}_l = \sum_{i=1}^{k^2} W_i \odot X_i,
\end{equation} 
\begin{equation}
	\mathcal{F}_h = \hat{F}_{LL}-\mathcal{F}_{l},
\end{equation}where $\odot$ is element-wise multiplication, $i$ denotes index of each spatial element in the $k^2$ patch. Hence, the FSB facilitates content-adaptive frequency decomposition by learning data-driven low-pass filtering, in contrast to conventional wavelet transforms that rely on fixed filter kernels, thereby achieving more accurate structural preservation and enhanced texture recovery.

\noindent\textbf{Low-Frequency Modulation.}  
As shown in Fig.~\ref{fig:fft}(c), the obtained low-frequency representation still contains residual mid-frequency structures. To further refine its spectral purity, we perform amplitude-phase fusion in the Fourier domain, where structural cues (phase) and global intensity trends (amplitude) can be adaptively controlled, yielding cleaner and more expressive low-frequency features. The whole process can be formulated as follows:
\begin{equation}
\begin{split}
\tilde L &= \operatorname{IFFT}\Big(
  \mathcal{F}_{\mathrm{conv1}}\big( \operatorname{Concat}(\mathcal{A}_{\hat F_{LL}}, \mathcal{A}_{\mathcal{F}_l})\big),\\
&\qquad
  \mathcal{F}_{\mathrm{conv2}}\big( \operatorname{Concat}(\Phi_{\hat F_{LL}}, \Phi_{\mathcal{F}_l})\big)
\Big),
\end{split}
\end{equation}
where IFFT(,) denotes Inverse Fast Fourier Transform, Concat(,) means channel concatenation, and $\mathcal{A}(\cdot)$ and $\Phi(\cdot)$ denote the amplitude and phase components, respectively. $\mathcal{F}_{\text{conv}_1}(\cdot)$ and $\mathcal{F}_{\text{conv}_2}(\cdot)$ represent adaptive fusion operations applied to the amplitude and phase in the spectral domain. This spectral redistribution transforms static DWT decomposition into a learnable low-frequency enhancement 

\newcommand{\graytext}[1]{\textcolor{black!60}{#1}}
\begin{table*}[t]
  \centering  
  \caption{Single-task restoration results, including denoising on Kodak24, dehazing on SOTS, deraining on Rain100L, deblurring on GoPro, and low-light enhancement on LOL. 
  \textbf{Bold} and \underline{underline} denote the best and the second best of AiOIR methods.}
  \vspace{-1em}
  \renewcommand\arraystretch{0.6}
  \tabcolsep=2pt
  \resizebox{0.95\linewidth}{!}{%
  \begin{tabular}{lccc|lc|lc|lc|lc}
    \toprule[1pt]
    {\multirow{2}{*}{\textbf{Kodak24}}}
      & \multicolumn{3}{c|}{\textbf{PSNR}} 
      & {\multirow{2}{*}{\textbf{SOTS}}}  &  \multirow{2}{*}{\textbf{PSNR/SSIM}}
      & {\multirow{2}{*}{\textbf{Rain100L}}}  & \multirow{2}{*}{\textbf{PSNR/SSIM}}
      & {\multirow{2}{*}{\textbf{GoPro}}}  & \multirow{2}{*}{\textbf{PSNR/SSIM}}
      & {\multirow{2}{*}{\textbf{LOL}}}  & \multirow{2}{*}{\textbf{PSNR/SSIM}} \\
      \cline{2-4}
      & $\sigma=15$ & $\sigma=25$ & $\sigma=50$
      &            &  
      &            &  
      &            &  
      &            &  \\
    \midrule 
    \graytext{DnCNN~\cite{DnCNN}} & \graytext{34.60} & \graytext{32.14} & \graytext{28.95} &
    \graytext{DehazeNet~\cite{DehazeNet}} & \graytext{22.46/0.851} &
    \graytext{UMR~\cite{UMR}} & \graytext{32.39/0.921} &
    \graytext{DeblurGAN~\cite{DeblurGAN}} & \graytext{28.70/0.858} &
    \graytext{URetinex~\cite{URetinex}} & \graytext{21.33/0.835} \\
    
    \graytext{FFDNet~\cite{FFDNet}} & \graytext{34.63} & \graytext{32.13} & \graytext{28.98} &
    \graytext{FDGAN~\cite{FDGAN}} & \graytext{23.15/0.921} &
    \graytext{LPNet~\cite{LPNet}} & \graytext{33.61/0.958} &
    \graytext{Stripformer~\cite{Stripformer}} & \graytext{33.08/0.962} &
    \graytext{SMG~\cite{SMG}} & \graytext{23.81/0.809} \\
    
    \graytext{ADFNet~\cite{ADFNet}} & \graytext{34.77} & \graytext{32.22} & \graytext{29.06} &
    \graytext{DehazeFormer~\cite{DehazeFormer}} & \graytext{31.78/0.977} &
    \graytext{DRSformer~\cite{DRSformer}} & \graytext{38.14/0.983} &
    \graytext{HI-Diff~\cite{HI-Diff}} & \graytext{33.33/0.964} &
    \graytext{Retinexformer~\cite{Retinexformer}} & \graytext{25.16/0.845} \\
    
    \graytext{MIRNet-v2~\cite{MIRNet_v2}} & \graytext{34.29} & \graytext{31.81} & \graytext{28.55} &
    \graytext{Restormer~\cite{Restormer}} & \graytext{30.87/0.969} &
    \graytext{Restormer~\cite{Restormer}} & \graytext{36.74/0.978} &
    \graytext{MPRNet~\cite{MPRNet}} & \graytext{32.66/0.959} &
    \graytext{MIRNet~\cite{MIRNet}} & \graytext{24.14/0.835} \\
    
    \graytext{Restormer~\cite{Restormer}} & \graytext{34.78} & \graytext{32.37} & \graytext{29.08} &
    \graytext{NAFNet~\cite{NAFNet}} & \graytext{30.98/0.970} &
    \graytext{NAFNet~\cite{NAFNet}} & \graytext{36.63/0.977} &
    \graytext{Restormer~\cite{Restormer}} & \graytext{32.92/0.961} &
    \graytext{Restormer~\cite{Restormer}} & \graytext{22.43/0.823} \\
    
    \graytext{NAFNet~\cite{NAFNet}} & \graytext{34.27} & \graytext{31.80} & \graytext{28.62} &
    \graytext{FSNet~\cite{FSNet}} & \graytext{31.11/0.971} &
    \graytext{FSNet~\cite{FSNet}} & \graytext{37.27/0.980} &
    \graytext{FSNet~\cite{FSNet}} & \graytext{33.29/0.963} &
    \graytext{DiffIR~\cite{DiffIR}} & \graytext{23.15/0.828} \\

    \hline
    \hline
    AirNet~\cite{AirNet}            & 34.81 & 32.44 & 29.10  
      & AirNet~\cite{AirNet}        & 23.18/0.900 
      & AirNet~\cite{AirNet}        & 34.90/0.977 
      & AirNet~\cite{AirNet}        & 31.64/0.945 
      & AirNet~\cite{AirNet}        & 21.52/0.832 \\
    IDR~\cite{IDR}               & 34.78 & 32.42 & 29.13  
      & PromptIR~\cite{PromptIR}      & 31.31/0.973 
      & PromptIR~\cite{PromptIR}      & 37.04/0.979 
      & PromptIR~\cite{PromptIR}      & \underline{32.41}/\underline{0.956} 
      & PromptIR~\cite{PromptIR}      & 22.97/0.834 \\
    Perceive-IR~\cite{DBLP:journals/corr/abs-2408-15994}       & \underline{34.84} & \underline{32.50} & \underline{29.16}  
      & Perceive-IR~\cite{DBLP:journals/corr/abs-2408-15994}   & \underline{31.65}/\underline{0.977} 
      & Perceive-IR~\cite{DBLP:journals/corr/abs-2408-15994}   & \textbf{38.41/0.984} 
      & Perceive-IR~\cite{DBLP:journals/corr/abs-2408-15994}   & \textbf{32.83}/\textbf{0.960} 
      & Perceive-IR~\cite{DBLP:journals/corr/abs-2408-15994}   & \underline{23.79}/\underline{0.841} \\
    \rowcolor{my_color}\textbf{ClusIR (Ours)}
                      & \textbf{35.06} & \textbf{32.60} & \textbf{29.49}  
      & \textbf{ClusIR (Ours)} & \textbf{32.67/0.983} 
      & \textbf{ClusIR (Ours)} & \underline{37.52}/\underline{0.980} 
      & \textbf{ClusIR (Ours)} & 30.07/0.904 
      & \textbf{ClusIR (Ours)} & \textbf{23.82}/\textbf{0.852}        \\
    \bottomrule[1pt]
  \end{tabular}}
  \label{tab:single_task_all}
  % \vspace{-1em}
\end{table*}

\begin{table*}[t]
\caption{Comparison to state-of-the-art on composited degradations. 
PSNR (dB) and SSIM are reported on full RGB images.}
\vspace{-1em}
\centering
\setlength{\tabcolsep}{1pt}
\renewcommand\arraystretch{0.6}
\resizebox{0.95\linewidth}{!}{
\begin{tabular}{l|cccc|ccccc|cc|c}
\toprule
\multirow{2}{*}{\textbf{Method}} &
\multicolumn{4}{c|}{\textbf{CDD11-Single}} &
\multicolumn{5}{c|}{\textbf{CDD11-Double}} &
\multicolumn{2}{c|}{\textbf{CDD11-Triple}} &
\multirow{2}{*}{\textbf{Avg.}} \\
\cline{2-12}
& L & H & R & S & L+H & L+R & L+S & H+R & H+S & L+H+R & L+H+S & \\ 
\midrule

AirNet~\cite{AirNet}
& 24.83/0.778 & 24.21/0.951 & 26.55/0.891 & 26.79/0.919 
& 23.23/0.779 & 22.82/0.710 & 23.29/0.723 & 22.21/0.868 & 23.29/0.901 
& 21.80/0.708 & 22.24/0.725 
& 23.75/0.814 \\

PromptIR~\cite{PromptIR}
& \underline{26.32}/\underline{0.805} & 26.10/0.969 & 31.56/0.946 & 31.53/0.960
& \underline{24.49}/0.789 & 25.05/0.771 & 24.51/0.761 & 24.54/0.924 & 27.05/0.925 
& 23.74/0.752 & 23.33/0.747 
& 25.90/0.850 \\

WGWSNet~\cite{WGWSNet}
& 24.39/0.774 & \underline{27.90}/\textbf{0.982} & \textbf{33.15}/\textbf{0.964} & \textbf{34.43}/\underline{0.973}
& 24.27/\underline{0.800} & \underline{25.06}/\underline{0.772} & \underline{24.60}/\underline{0.765} & \underline{27.23}/\textbf{0.955} & \textbf{27.65}/\textbf{0.960}
& \underline{23.90}/\underline{0.772} & \underline{23.97}/\underline{0.771}
& \underline{26.96}/\underline{0.863} \\

WeatherDiff~\cite{WeatherDiff}
& 23.58/0.763 & 21.99/0.904 & 24.85/0.885 & 24.80/0.888
& 21.83/0.756 & 22.69/0.730 & 22.12/0.707 & 21.25/0.868 & 21.99/0.868
& 21.23/0.716 & 21.04/0.698
& 22.49/0.799 \\

\rowcolor{my_color}\textbf{ClusIR (Ours)}
& \textbf{26.62}/\textbf{0.825} & \textbf{28.58}/\textbf{0.982} & \underline{32.59}/\textbf{0.964} & \underline{33.52}/\textbf{0.975}
& \textbf{24.87}/\textbf{0.814} & \textbf{25.60}/\textbf{0.799} & \textbf{25.41}/\textbf{0.795} & \textbf{27.52}/\textbf{0.955} & \underline{27.48}/\textbf{0.960}
& \textbf{24.31}/\textbf{0.780} & \textbf{24.18}/\textbf{0.782}
& \textbf{27.33}/\textbf{0.878} \\
\bottomrule
\end{tabular}}
\label{cdd}
% \vspace{-1em}
\end{table*}

\noindent\textbf{High-Frequency Modulation.} We also use the self-mined prior $\mathcal{F}_h$ to further modulate the  high-frequency subbands. Specifically, high-frequency subbands $\hat{F}_{HL}, \hat{F}_{LH}, \hat{F}_{HH}$ are first concatenated and modulated by $\mathcal{F}_h$ through channel-wise gating, and this process can be defined as follows:
\begin{equation}
	\tilde{H} = \sigma(\mathbf{W} \mathcal{F}_h) \cdot [\hat{F}_{HL}, \hat{F}_{LH}, \hat{F}_{HH}],
\end{equation}where $\mathbf{W}$ denotes a learnable projection. As shown in Fig.~\ref{fig:fft}(f), this modulation adaptively emphasizes salient edges and textures, enabling the model to capture fine-grained high-frequency structures with enhanced spectral fidelity. Finally, the enhanced low- and high-frequency features are progressively combined through inverse wavelet reconstruction, yielding the restored representation. Therefore, by incorporating semantic prompts with FSB, DAFMM effectively enhances both structural integrity and fine-grained texture details, thereby achieving more effective image restoration.

\section{Experiments}
\label{sec:Experiment}
\subsection{Experimental Setup}
% \noindent\textbf{Datasets.} Following~\cite{AirNet, PromptIR, DBLP:journals/corr/abs-2408-15994}, we consider two training paradigms: \textit{One-by-One} (single-task training) and \textit{All-in-One} (multi-task joint training). Specifically, we explore two common degradation combinations for AiOIR: {\textbf{N}}+{\textbf{H}}+{\textbf{R}} (Noise, Haze, Rain) and {\textbf{N}}+{\textbf{H}}+{\textbf{R}}+{\textbf{B}}+{\textbf{L}} (Noise, Haze, Rain, Blur, Low-light). We summarize all datasets used as follows. 
% \textbf{Denoising}: BSD400~\cite{BSD400} (400) + WED~\cite{WED} (4,744) for training with Gaussian noise ($\sigma \in \{15, 25, 50\}$); tested on CBSD68~\cite{BSD68} and Kodak24~\cite{Kodak24}.
%   \textbf{Dehazing}: OTS (72,135 pairs) from RESIDE-$\beta$~\cite{SOTS} for training; tested on SOTS-Outdoor~\cite{SOTS} (500).
%    \textbf{Deraining}: Rain100L~\cite{Rain100L}, with 200/100 training/testing pairs.
%   \textbf{Deblurring}: GoPro~\cite{GoPro}, with 2,103/1,111 training/testing pairs.
%   \textbf{Low-light}: LOL~\cite{LOL}, with 485/15 training/testing pairs.
\noindent\textbf{Datasets.}
% , and \textit{multi-degradation combination training} that involves mixed or composited degradations
Following~\cite{AirNet, PromptIR, DBLP:journals/corr/abs-2408-15994}, we use three common training paradigms: \textit{One-by-One} (single-task training), \textit{All-in-One} (multi-task joint training), and \textit{multi-degradation combination training}. For image denoising task, we combine the BSD400~\cite{BSD400} and WED~\cite{WED} datasets, adding Gaussian noise with levels $\sigma \in {\{15, 25, 50}\}$, and evaluate on CBSD68~\cite{BSD68} and Kodak24~\cite{Kodak24}. The dehazing task uses the RESIDE-$\beta$~\cite{SOTS} dataset, while deraining adopts Rain100L~\cite{Rain100L}. For deblurring and low-light enhancement, we employ the GoPro~\cite{GoPro} and LOL~\cite{LOL} datasets, respectively.
To develop a unified restoration model, we merge these datasets in a three (\textcolor{blue}{\bf{N}}+\textcolor{green}{\bf{H}}+\textcolor{red}{\bf{R}}) or five (\textcolor{blue}{\textbf{N}}+\textcolor{green}{\textbf{H}}+\textcolor{red}{\bf{R}}+\textcolor{purple}{\textbf{B}}+\textcolor{pink}{\textbf{L}}) degradation setting. For \textit{multi-degradation combination training}, we use the CDD11 dataset~\cite{OneRestore}, where multiple degradation types are jointly applied to a single image to simulate more realistic and challenging scenarios. For \textit{real-world combination training}, we use the WeatherBench dataset\cite{weatherbench}, which contains complex real-world environments, providing a more practical evaluation of model generalization.

% For the composite degradation setting, we use the CDD11 dataset~\cite{OneRestore}, where multiple degradation types are jointly applied to a single image to simulate more realistic and challenging scenarios.

\noindent\textbf{Implementation Details.}
For ClusIR, we employ the Adam optimizer ($\beta_1{=}0.9$, $\beta_2{=}0.999$) with an initial learning rate of $2{\times}10^{-4}$ and a joint objective combining $\ell_1$ and MS-SSIM losses~\cite{MSSSIM}. The model is trained for 150 epochs with a batch size of 48, and the learning rate is halved after 75 epochs. Following~\cite{DBLP:conf/iclr/0001ZKKSK25}, we set the data expansion ratios to 3, 120, 5, and 200 for denoising, deraining, deblurring, and low-light enhancement, respectively, while keeping dehazing unchanged. We set the cluster prototypes and activated clusters to [3, 3, 3, 3] and [2, 2, 2, 2], respectively, across all stages. Training is conducted on 4 NVIDIA A100 GPUs using random cropping and flipping with $128{\times}128$ patches for data augmentation.

\subsection{All-in-One Restoration Results}
% We evaluate ClusIR under two \textit{All-in-One} configurations with progressively increasing complexity: a moderate three-degradation setting (\textcolor{blue}{\bf{N}}+\textcolor{green}{\bf{H}}+\textcolor{red}{\bf{R}}), a comprehensive five-degradation setting (\textcolor{blue}{\textbf{N}}+\textcolor{green}{\textbf{H}}+\textcolor{red}{\bf{R}}+\textcolor{purple}{\textbf{B}}+ \textcolor{pink}{\textbf{L}}).
We evaluate ClusIR under two \textit{All-in-One} settings: a three-degradation setting (\textcolor{blue}{\bf N}+\textcolor{green}{\bf H}+\textcolor{red}{\bf R}) 
and a five-degradation setting (\textcolor{blue}{\bf N}+\textcolor{green}{\bf H}+\textcolor{red}{\bf R}+\textcolor{purple}{\bf B}+\textcolor{pink}{\bf L}).

% We evaluate ClusIR under five settings with increasing complexity: 
% (i) a three-degradation setting (\textcolor{blue}{\bf N}+\textcolor{green}{\bf H}+\textcolor{red}{\bf R}), 
% (ii) a five-degradation setting (\textcolor{blue}{\bf N}+\textcolor{green}{\bf H}+\textcolor{red}{\bf R}+\textcolor{purple}{\bf B}+\textcolor{pink}{\bf L}),  
% (iii) an One-by-One setting,  
% (iv) a composited setting with multiple co-occurring degradations, 
% and (v) real-world scenarios setting.
\begin{figure}[t]
  \centering
  \setlength{\abovecaptionskip}{2pt}   % 图与caption间距
  \setlength{\belowcaptionskip}{0pt}   % caption与正文间距
  \includegraphics[width=\linewidth]{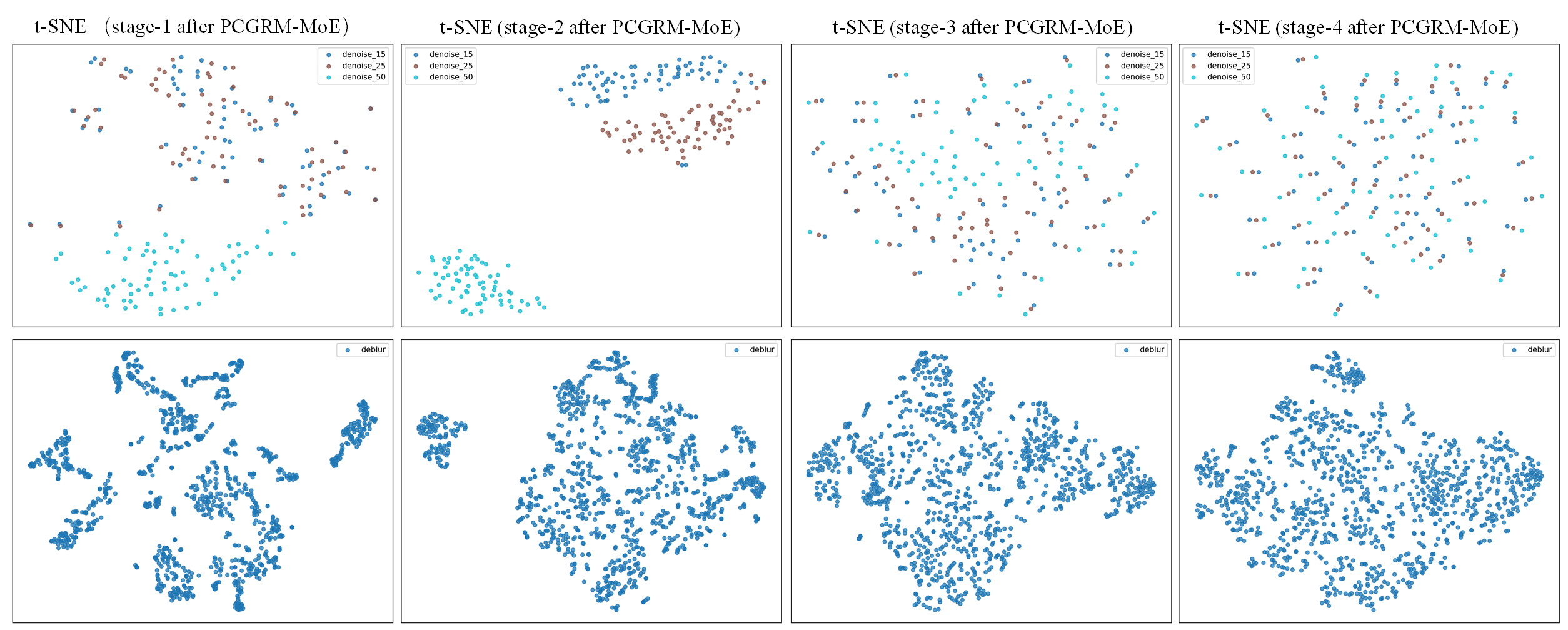}
  \caption{
  t-SNE visualization of cross-stage feature distributions after the PCGRM-MoE under the One-by-One setting.
  }
  \label{fig:tsne_onebyone}
  \vspace{-1em}
\end{figure}

\noindent\textbf{Results Comparisons on Three Tasks:} Tab.~\ref{tab:three_task} presents the quantitative comparison under the All-in-One (``\textcolor{blue}{\bf{N}}+\textcolor{green}{\bf{H}}+\textcolor{red}{\bf{R}}'') setting. ClusIR achieves the best overall performance with an average PSNR of 33.06 dB and SSIM of 0.923, surpassing all recent All-in-One methods. Compared with AdaIR, MoCE-IR, and DFPIR, ClusIR yields gains of +0.37 dB/+0.005, +0.33 dB/+0.006, and +0.18 dB/+0.004 in average PSNR/SSIM, respectively. Notably, ClusIR achieves new state-of-the-art results on dehazing (32.85/0.983) and deraining (38.71/0.984), while maintaining competitive denoising accuracy across all noise levels. These improvements demonstrate that the proposed PCGRM and DAFMM jointly enhance cross-degradation generalization and structural fidelity in complex restoration scenarios.
\begin{table*}[t]
\caption{Quantitative comparison on the real-world WeatherBench dataset. \textbf{Bold} and \underline{underline} denote the best and the second best of AiOIR methods.}
\vspace{-0.8em}
\centering
\renewcommand\arraystretch{0.6}
\resizebox{0.95\linewidth}{!}{
\begin{tabular}{l|c|ccc|ccc|ccc|ccc}
\hline
\multirow{2}{*}{\textbf{Method}} & \multirow{2}{*}{\textbf{Venue}} 
& \multicolumn{3}{c|}{\textbf{Dehaze}} 
& \multicolumn{3}{c|}{\textbf{Derain}} 
& \multicolumn{3}{c|}{\textbf{Desnow}} 
& \multicolumn{3}{c}{\textbf{Avg.}} \\
\cline{3-14}
& & PSNR$\uparrow$ & SSIM$\uparrow$ & LPIPS$\downarrow$
  & PSNR$\uparrow$ & SSIM$\uparrow$ & LPIPS$\downarrow$
  & PSNR$\uparrow$ & SSIM$\uparrow$ & LPIPS$\downarrow$
  & PSNR$\uparrow$ & SSIM$\uparrow$ & LPIPS$\downarrow$ \\
\hline
% Restormer\cite{Restormer} & CVPR'22 
% & 19.30 & 0.687 & 0.412 
% & 34.49 & 0.945 & 0.197
% & 27.95 & 0.836 & 0.197
% & 27.25 & 0.823 & 0.269 \\

% DRSformer\cite{DRSformer} & CVPR'23 
% & 19.95 & 0.694 & 0.404 
% & 33.98 & 0.943 & 0.209 
% & 28.00 & 0.836 & 0.197 
% & 27.31 & 0.824 & 0.270 \\

% DCMPNet\cite{DCMPNet} & CVPR'24 
% & 21.18 & 0.506 & 0.491 
% & 32.04 & 0.876 & 0.282 
% & 24.81 & 0.614 & 0.546 
% & 26.01 & 0.665 & 0.440 \\
\graytext{Restormer\cite{Restormer}} & \graytext{CVPR'22} 
& \graytext{19.30} & \graytext{0.687} & \graytext{0.412} 
& \graytext{34.49} & \graytext{0.945} & \graytext{0.197}
& \graytext{27.95} & \graytext{0.836} & \graytext{0.197}
& \graytext{27.25} & \graytext{0.823} & \graytext{0.269} \\

\graytext{DRSformer\cite{DRSformer}} & \graytext{CVPR'23} 
& \graytext{19.95} & \graytext{0.694} & \graytext{0.404} 
& \graytext{33.98} & \graytext{0.943} & \graytext{0.209} 
& \graytext{28.00} & \graytext{0.836} & \graytext{0.197} 
& \graytext{27.31} & \graytext{0.824} & \graytext{0.270} \\

\graytext{DCMPNet\cite{DCMPNet}} & \graytext{CVPR'24} 
& \graytext{21.18} & \graytext{0.506} & \graytext{0.491} 
& \graytext{32.04} & \graytext{0.876} & \graytext{0.282} 
& \graytext{24.81} & \graytext{0.614} & \graytext{0.546} 
& \graytext{26.01} & \graytext{0.665} & \graytext{0.440} \\

\hline\hline

AirNet\cite{AirNet} & CVPR'22 
& 20.94 & 0.705 & \underline{0.383}
& 33.59 & 0.942 & 0.224 
& 22.06 & 0.780 & 0.291 
& 25.53 & 0.809 & \underline{0.299} \\

TransWeather\cite{Transweather} & CVPR'22 
& 19.79 & 0.680 & 0.397 
& 29.34 & 0.903 & 0.294 
& 24.96 & 0.796 & 0.231 
& 24.70 & 0.793 & 0.307 \\

WGWSNet\cite{WGWSNet} & CVPR'23 
& 13.79 & 0.603 & 0.535 
& \textbf{37.08} & \textbf{0.961} & \textbf{0.117} 
& 20.81 & 0.780 & 0.248 
& 23.89 & 0.781 & 0.300 \\

PromptIR\cite{PromptIR} & NeurIPS'23 
& \underline{21.11} & \textbf{0.713} & \textbf{0.375} 
& 34.54 & \underline{0.944} & \underline{0.198} 
& \textbf{27.93} & \textbf{0.836} & \textbf{0.195} 
& \underline{27.86} & \textbf{0.831} & \textbf{0.256} \\

Histoformer\cite{Histoformer} & ECCV'24 
& 17.69 & 0.669 & 0.437 
& 30.70 & 0.916 & 0.279 
& 25.39 & 0.808 & \underline{0.225 }
& 24.59 & 0.798 & 0.314 \\

\hline

\rowcolor{my_color}\textbf{ClusIR} & Ours 
& \textbf{22.92} & \underline{0.704} & \underline{0.383}
& \underline{35.00} & 0.938 & 0.258 
& \underline{27.75} & \underline{0.821} & 0.258 
& \textbf{28.56} & \underline{0.821} & \underline{0.299} \\
\hline

\end{tabular}
}
\vspace{-1em}
\label{tab:weatherbench_comparison}
\end{table*}

\begin{figure*}[t]
  \centering
  \includegraphics[width=0.95\linewidth]{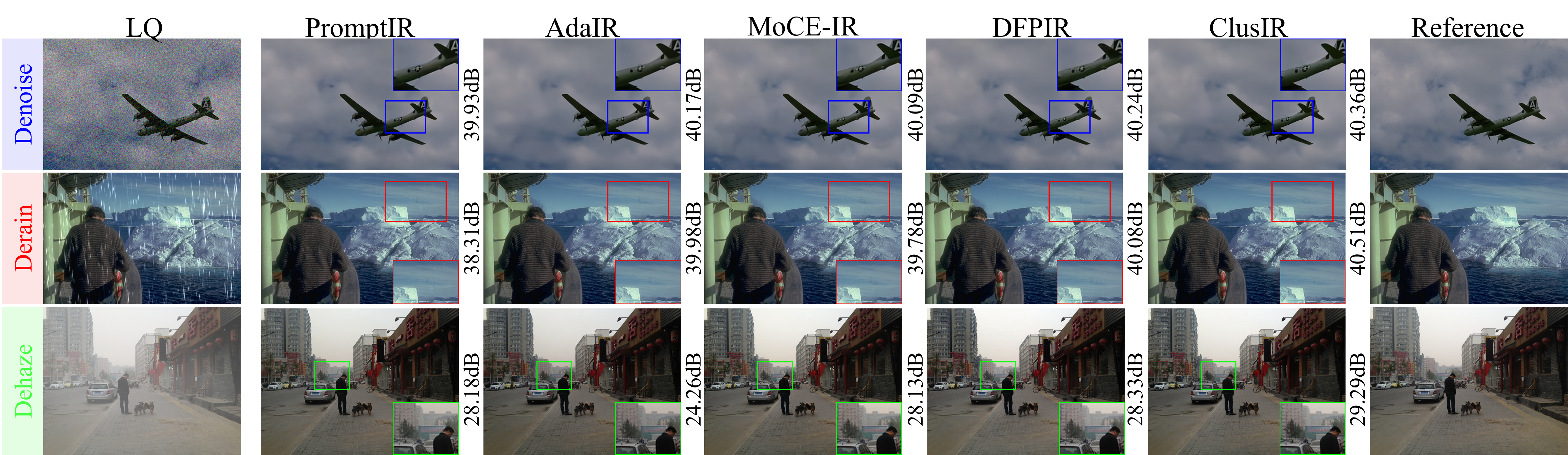}
      \vspace{-1em}
  \caption{
  Visual comparison of ClusIR with SOTA AiOIR methods under the “\textcolor{blue}{\bf{N}}+\textcolor{green}{\bf{H}}+\textcolor{red}{\bf{R}}” setting.
  More visualization results are provided in the supplementary materials.
  }
  \label{fig:3d}
  \vspace{-1em}
\end{figure*}
\noindent\textbf{Results Comparisons on Five Tasks:} Tab.~\ref{tab:five_task} summarizes the results under this more challenging five-degradation setting. ClusIR achieves the second-highest PSNR (30.58 dB) and the best SSIM (0.919) among all competing All-in-One methods. Notably, ClusIR delivers leading performance in dehazing, outperforming AdaIR, MoCE-IR, and DFPIR with a score of 31.94/0.982, and also achieves clear gains in low-light enhancement, attaining +0.015 and +0.024 SSIM over AdaIR and DFPIR, respectively.
For denoising, deraining, and deblurring, ClusIR maintains competitive performance, reaching 31.35/0.894, 37.28/0.980, and 28.54/0.872, respectively, which demonstrates its stable restoration capability across the remaining tasks.

\subsection{One-by-One Restoration Results}
We evaluate ClusIR under the \textit{One-by-One} setting (Tab.~\ref{tab:single_task_all}), where each restoration task is trained and tested independently. Top lines are task-specific and general methods, and the bottom lines refer to AiOIR methods.

As shown in Tab.~\ref{tab:single_task_all}, ClusIR achieves the best denoising performance on Kodak24, consistently ranking first in all noise levels, Notably, it reaches  35.06~dB at $\sigma=15$, surpassing Perceive-IR and Restormer by +0.22~dB and +0.28~dB, respectively. On the dehazing task, ClusIR leads on SOTS, attaining 32.67/0.983 on SOTS, outperforming DehazeFormer and Perceive-IR. For deraining, it achieves 37.52/0.980 on Rain100L. On the low-light enhancement task, ClusIR yields 23.82/0.852, ranking first in AiOIR methods. For deblurring, HI-Diff and FSNet lead the performance due to their dedicated blur-handling designs.

Building on these results, we further examine the feature distributions in the single‑task scenario. As illustrated in Fig.~\ref{fig:tsne_onebyone}, ClusIR fails to effectively activate the cluster prototypes, which undermines the PCGRM, prevents reliable degradation discrimination, and consequently limits the generation of semantic prompts. In contrast, for the denoising task, ClusIR treats different Gaussian noise levels as distinct degradation types during joint training, allowing cluster semantics to emerge progressively across stages. Specifically, at stage~1, the separability among noise levels remains ambiguous; by stage~2, noticeable degradation discrimination begins to appear; and by stage~4, the features of the same sample across $\sigma{=}15/25/50$ become closer to each other after being refined by the PCGRM-MoE, while the features of different samples become more widely separated. These visual results further demonstrate the effectiveness of ClusIR in handling diverse degradation types.

% \begin{table}[t]
% \caption{LPIPS comparison under AiOIR-3 setting.}
% \vspace{-1.2em}
% \centering
% \renewcommand\arraystretch{0.1}
% \resizebox{\linewidth}{!}{
% \begin{tabular}{l|ccc|c|c|c}
% \toprule[1pt]
%  \multirow{2}{*}{\textbf{Method}} 
% & \multicolumn{3}{c|}{\textbf{Denoising} (CBSD68)} 
% & \multicolumn{1}{c|}{\textbf{Dehazing}} 
% & \multicolumn{1}{c|}{\textbf{Deraining}} 
% & \multirow{2}{*}{\textbf{Avg.}} \\
% \cmidrule(lr){2-6}
% & 
% $\,\sigma\!=\!15$ & $\,\sigma\!=\!25$ & $\,\sigma\!=\!50$ 
% & \multicolumn{1}{c|}{SOTS} 
% & Rain100L & \\
% \midrule
% PromptIR 
% & 0.1571 & 0.2216 & 0.3174 & 0.0456 & 0.0329 & 0.1549 \\

% MoCE-IR 
% & 0.1412 & 0.1996 & 0.2927 & 0.0338 & 0.0270 & 0.1380 \\

% % \rowcolor{my_color}
% \textbf{ClusIR} 
% & \textbf{0.1401} & \textbf{0.1932} & \textbf{0.2799} & \textbf{0.0286} & \textbf{0.0240} & \textbf{0.1332} \\
% \bottomrule[1pt]
% \end{tabular}}
% \label{tab:lpips_three_task}
% \vspace{-1.5em}
% \end{table}

\begin{figure}[t]
    \centering
    % \vspace{-1.3em}
    \includegraphics[width=0.95\linewidth]{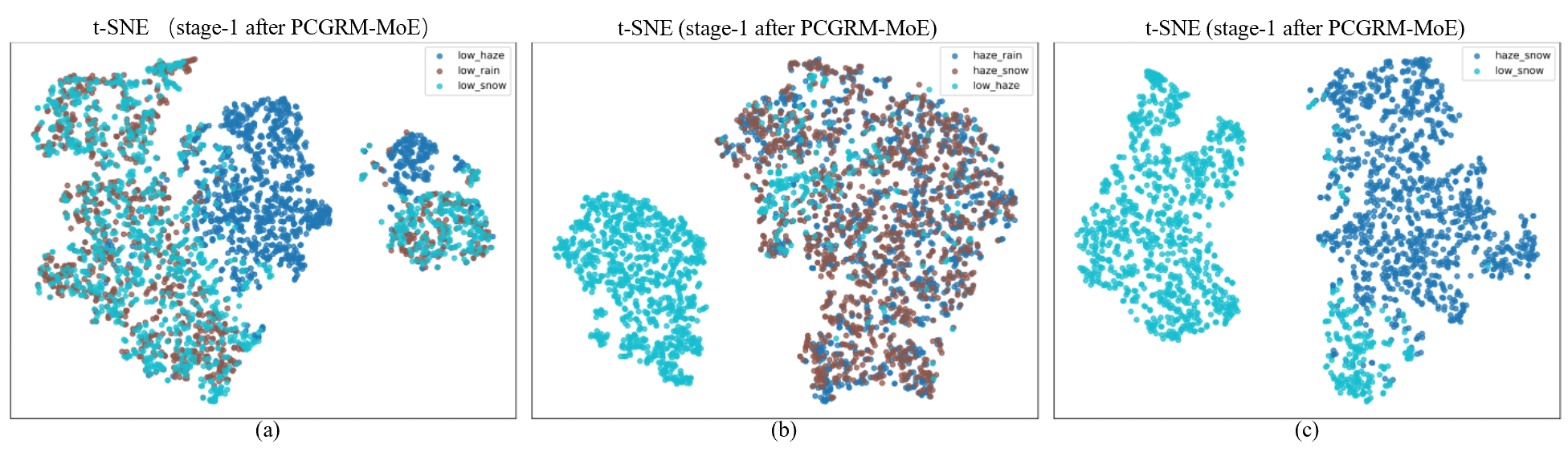}
    \vspace{-1em}
    % , demonstrating that ClusIR maintains strong feature disentanglement capability under mixed degradation scenarios.
    \caption{t-SNE analysis under composite degradations on stage1.}
    \label{fig:cddtsne}
     \vspace{-2em}
\end{figure}

\subsection{Composited Degradation Restoration Results}
Tab.~\ref{cdd} presents a comparison of ClusIR with state-of-the-art methods on the CDD11 dataset under Composited Degradation setting. ClusIR outperforms existing methods in terms of both PSNR and SSIM across all degradation combinations. Specifically, ClusIR achieves the highest performance, with an average PSNR of 27.33 dB and SSIM of 0.878, surpassing AirNet~\cite{AirNet}, PromptIR~\cite{PromptIR}, WGWSNet~\cite{WGWSNet}, and WeatherDiff~\cite{WeatherDiff} by gains of +3.58 dB / +0.075, +1.43 dB / +0.028, +0.37 dB / +0.015, and +4.84 dB / +0.079, respectively.
Notably, for the CDD11-Double and CDD11-Triple settings, ClusIR consistently delivers superior performance, highlighting its robustness in handling multiple types of image degradation. 
Overall, these results demonstrate ClusIR's effectiveness in all-in-one image restoration, showcasing its ability to simultaneously address multiple degradation tasks with a single unified model. These improvements can be attributed to our proposed PCGRM, which not only enables effective degradation discrimination but also activates the corresponding expert groups, facilitating the integration of specialized knowledge to handle diverse degradation types efficiently.

Moreover, Under composite degradations that share a low-light component (e.g., low+haze, low+snow, and low+rain) on the CDD11 dataset, which is designed to reflect complex and highly coupled real-world degradations, ClusIR still exhibits structured and interpretable clustering behavior (Fig.~\ref{fig:cddtsne}(a)). Specifically, ClusIR distinguishes haze from rain and snow, consistent with their frequency characteristics: haze is dominated by mid-to-low frequency components, while rain and snow exhibit stronger high-frequency patterns. Similar behavior is observed in (b) under the same haze condition and in (c) under snow degradation, where haze and low-light remain separable. Fig.~\ref{fig:cddtsne} indicates that the observed clustering patterns are not artifacts of isolated synthetic degradations, but reflect intrinsically discriminative representations learned by ClusIR under coupled and realistic degradation settings.

\subsection{Real-world Restoration Results}
To assess the robustness and generalization capability of the proposed ClusIR in real-world scenarios with unknown degradation mixtures, we conduct a more comprehensive experiments on the WeatherBench dataset\cite{weatherbench}. Top lines are task-specific and general methods, and the bottom lines refer to AiOIR methods.

Tab.~\ref{tab:weatherbench_comparison} reports the comparisons under real-world scenarios. ClusIR consistently achieves the best PSNR (28.56 dB), outperforming PromptIR\cite{PromptIR} and AirNet~\cite{AirNet} by +0.70 dB and +3.03 dB, respectively. Moreover, ClusIR attains the second-best SSIM (0.821) under the all-in-one paradigm compared to PromptIR~\cite{PromptIR}.
These results can be attributed to the proposed PCGRM, which explicitly disentangles degradation features, enabling ClusIR to handle real-world degradations beyond synthetic degradations. 

% As a result, ClusIR not only performs well on composite degradations but also maintains robust performance in complex real-world scenarios, with more diverse and strongly coupled degradation patterns.

\begin{figure}[t]
  \centering
  \setlength{\abovecaptionskip}{2pt}   % 图与caption间距
  \setlength{\belowcaptionskip}{0pt}   % caption与正文间距
  \includegraphics[width=\linewidth]{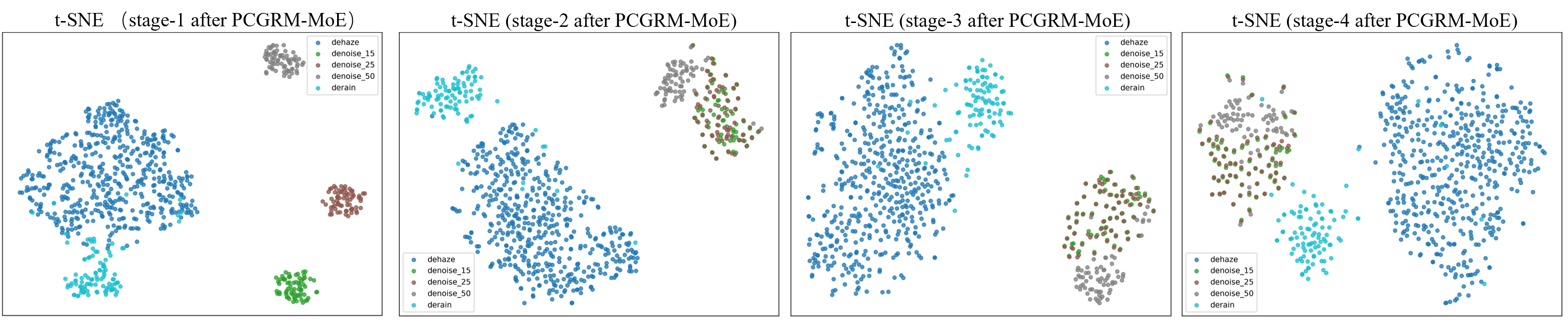}
  \caption{
  t-SNE visualization of cross-stage feature distributions after the PCGRM-MoE under the “\textcolor{blue}{\bf{N}}+\textcolor{green}{\bf{H}}+\textcolor{red}{\bf{R}}”  setting.
  }
  \label{fig:tsne}
  % \vspace{-1.5em}
\end{figure}

% Visual degradation of cluster prototype affinities under the AiOIR-3 setting.
\begin{figure}[t]
  \centering
  \setlength{\abovecaptionskip}{2pt}   % 图与caption间距
  \setlength{\belowcaptionskip}{0pt}   % caption与正文间距
  \includegraphics[width=\linewidth]{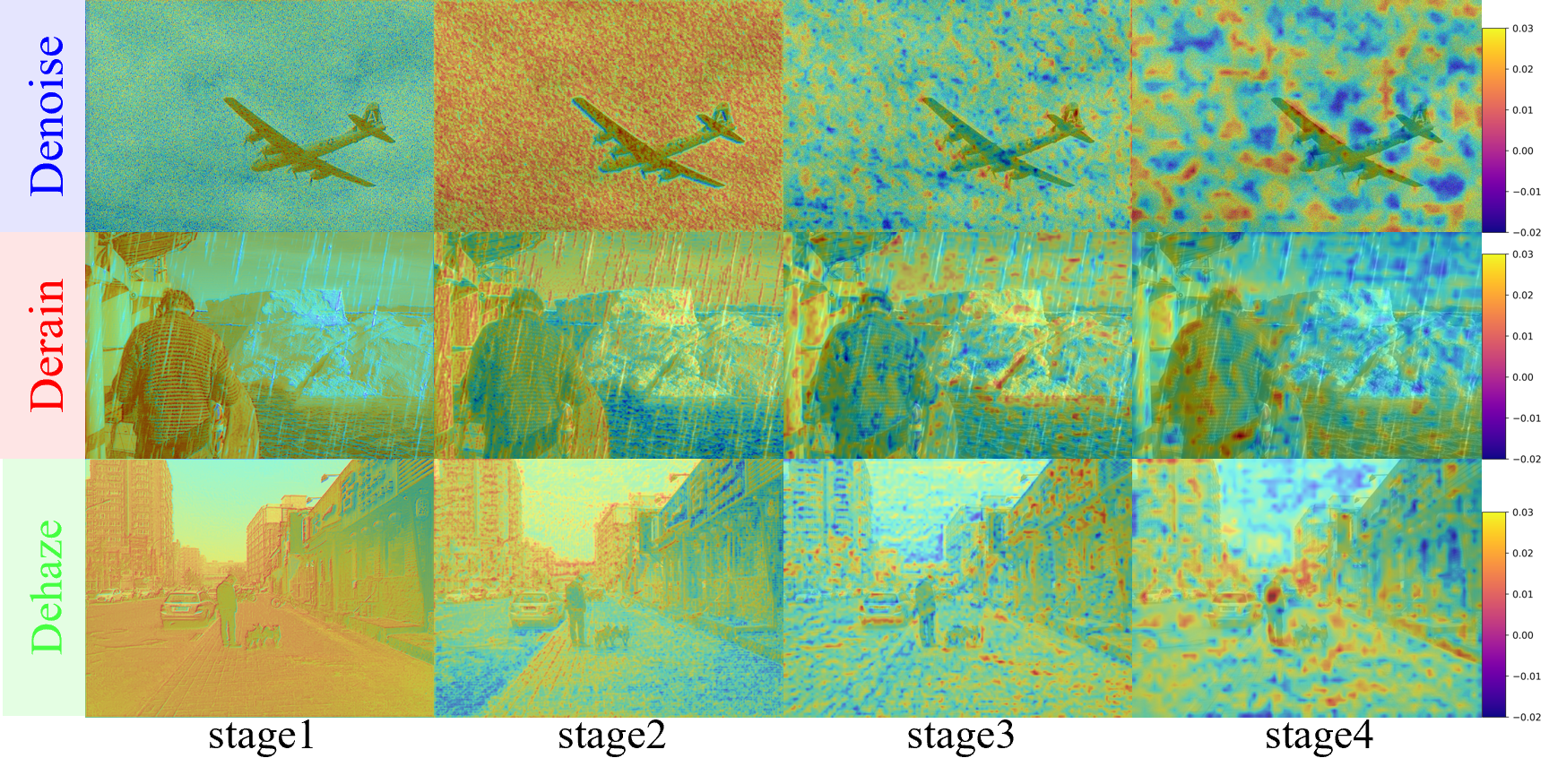}
  \vspace{-1.6em}
  \caption{Visual degradation of PCGRM-MoE under the “\textcolor{blue}{\bf{N}}+\textcolor{green}{\bf{H}}+\textcolor{red}{\bf{R}}” setting. Higher values mean stronger correlations.}
  \label{fig:qinheli}
  % \vspace{-2.5em}
\end{figure}

\begin{table}[!tb] 
  \centering
  \caption{Ablation study on the impact of key components. Each module contributes complementary gains.}
  \vspace{-1em}
  \renewcommand\arraystretch{0.6}
  \resizebox{\linewidth}{!}{%
    \begin{tabular}{c|ccc|cc}
      \toprule[1pt]
      Index
      & \multicolumn{1}{c}{WTB}
      & {PCGRM-MoE}
      & {DAFMM}
      & \textbf{avg PSNR}$\uparrow$
      & \textbf{avg SSIM}$\uparrow$ \\
      \midrule
      (a)
      & \ding{51}&  &   
      & 32.68 & 0.922 \\

      (b) 
      & \ding{51}& \ding{51} &   
      & 32.95 & 0.922 \\

      (c)
      & \ding{51}& \ding{51} & \ding{51}    
      & 33.06 & 0.923 \\

      \bottomrule[1pt]
    \end{tabular}%
  }
  \label{tab:alb_comp} 
  \vspace{-1.5em}
\end{table}

% \begin{table}[!tb]
%   \centering
%   \caption{Ablation on the impact of cluster prototypes across stages. 
%   We vary the number and activation of cluster prototypes in each stage.}
%   \vspace{-0.6em}
%   \renewcommand\arraystretch{1.4}
%   \resizebox{\linewidth}{!}{%
%   \begin{tabular}{c|c|c|cc}
%     \toprule[1pt]
%     Index 
%     & Cluster number (stage1–4)
%     & Activated clusters (stage1–4)
%     & \textbf{avg} PSNR 
%     & \textbf{avg} SSIM \\
%     \midrule
%     (1) & [3, 3, 3, 3] & [2, 2, 2, 2] & 33.06 & 0.923 \\
%     (2) & [2, 3, 4, 6] & [2, 2, 2, 2] & 32.27 & 0.912 \\
%     (3) & [6, 4, 3, 2] & [2, 2, 2, 2] & 32.40 & 0.912 \\
%     % (d) & [3, 3, 3, 3] & [1, 1, 1, 1] & 32.28 & 0.912 \\
%     \bottomrule[1pt]
%   \end{tabular}%
%   }
%   \label{tab:cluster_ablation}
%   \vspace{-1.3em}
% \end{table}

\begin{table}[!tb]
  \centering
  \caption{Ablation on the impact of cluster prototypes across stages. 
  We vary the number of cluster prototypes in each stage.}
  \vspace{-1em}
  \renewcommand\arraystretch{0.6}
  \resizebox{\linewidth}{!}{%
  \begin{tabular}{c|c|cc}
    \toprule[1pt]
    Index 
    & Cluster prototype numbers (stage1–4)
    & \textbf{avg PSNR}$\uparrow$
    & \textbf{avg SSIM}$\uparrow$ \\
    \midrule
    (1) & [3, 3, 3, 3] & 33.06 & 0.923 \\
    (2) & [2, 3, 4, 6] & 32.27 & 0.912 \\
    (3) & [6, 4, 3, 2] & 32.40 & 0.912 \\
    \bottomrule[1pt]
  \end{tabular}%
  }
  \label{tab:cluster_ablation}
  % \vspace{-2em}
\end{table}

\subsection{Visual Results}
As shown in Fig.~\ref{fig:3d}, ClusIR consistently improves texture fidelity and structural coherence across diverse degradations. In the red, blue, and green regions, it restores finer details, removes rain streaks more thoroughly, and reconstructs depth-consistent structures with balanced contrast, outperforming PromptIR, AirNet, and MoCE-IR. This improvement can be attributed to the strong degradation discrimination and semantic perception provided by the PCGRM. The t-SNE visualizations in Fig.~\ref{fig:tsne} further depict a coherent representational evolution, where diverse degradation features that are initially well separated progressively converge within related degradation families (e.g., denoise-15/25/50), while maintaining clear distinctions across different tasks (e.g., denoising, deraining, and dehazing). This reveals a  degradation-level discrimination to semantic alignment by cluster prototypes. 

Moreover, the stage-wise prototype affinity maps in Fig.~\ref{fig:qinheli} reveal a progressive shift from localized degradation activations to globally coherent semantic responses. In the early stages (e.g., stage~1 and stage~2), the activations are concentrated on task-specific degraded regions, reflecting explicit degradation-type discrimination. In contrast, deeper stages exhibit more diffuse and semantically structured responses, capturing abstract regularities of scene content. Collectively, these analyses confirm that ClusIR effectively bridges degradation-specific cues and global structural semantics through hierarchical, degradation-aware representations.

\subsection{Ablation Study}
We conduct the ablation studies to demonstrate the
effectiveness of our proposed ClusIR. The results are reported under the “\textcolor{blue}{\bf{N}}+\textcolor{green}{\bf{H}}+\textcolor{red}{\bf{R}}” setting. Due to the page limitation, more detailed results are attached in the supplementary materials.

\noindent\textbf{Impact of Key Components.} 
We progressively integrate the PCGRM-MoE and the DAFMM into the baseline. As shown in Tab.~\ref{tab:alb_comp}, the performance improves progressively from (a) to (c).
Using only the WTB in (a) provides basic restoration ability, yielding an average PSNR/SSIM of 32.68/0.922. Introducing PCGRM-MoE in (b) yields clear gains of +0.27 dB. Further incorporating DAFMM in (c) raises the average PSNR/SSIM to 33.06/0.923. These results can be attributed to PCGRM, which explicitly encodes degradation semantics and encourages more specialized expert behavior. Moreover, DAFMM further enhances both high- and low-frequency information through frequency self-mining.

% confirming that frequency-domain modulation complements cluster-guided spatial representation for more effective restoration.

\noindent\textbf{Impact of Cluster Prototype Numbers across Stages.}
To investigate the influence of the cluster configuration on the hierarchical representation, we vary the number of prototype clusters across stages. As shown in Tab.~\ref{tab:cluster_ablation}, a uniform cluster prototype configuration (1) achieves the best balance between discrimination and consistency (33.06/0.923), while the unbalanced design (2) and (3) both lead to performance degradation. This decline indicates the improper allocation of cluster prototypes: excessive prototypes in shallow stages cause over-segmentation of degradation cues and interfere with low-level features learning, whereas excessive prototypes in deeper stages result in semantic dispersion and inconsistency among degradation types.

% As shown in Tab.~\ref{tab:cluster_ablation}, a uniform cluster prototypes configuration (a) achieves the best balance between discrimination and consistency (33.06/0.923). An unbalanced design (c) slightly degrades performance, while fewer activated clusters (d) further reduce representation quality. These results suggest that balanced cluster allocation and sufficient activation are crucial for robust hierarchical representation learning. 

\section{Conclusion}
\label{sec:Conclusion}
In this paper, we present ClusIR, an All-in-One image restoration framework that leverages explicit degradation cues to unify spatial and frequency-domain information processing. The explicit degradation cues are learned via the proposed PCGRM, which incorporates a hierarchical two-stage process that first infers a degradation-aware cluster posterior and then performs cluster-conditional expert routing. Building on this, DAFMM utilizes cluster-guided priors for adaptive frequency modulation, enabling coordinated enhancement of structural and textural components. Extensive experiments on diverse benchmarks demonstrate that ClusIR delivers competitive performance across AiOIR tasks, validating the effectiveness of our cluster-guided spatial–frequency modeling paradigm.

% \section{Acknowledgments}

% \section{Appendices}

% \begin{acks}

% \end{acks}

\bibliographystyle{ACM-Reference-Format}
\bibliography{sample-base}

\clearpage
\appendix
% ============================================================
% Supplementary material for the arXiv version.
% This file is intended to be included AFTER \appendix in the
% main paper. Do not add \documentclass, \begin{document},
% \title, \author, or \maketitle here.
% ============================================================

% Start the supplementary material on a new page.  The optional
% argument of \twocolumn creates a full-width (single-column)
% title block, while the supplementary body remains two-column.
\clearpage
\twocolumn[
\begin{@twocolumnfalse}
    \begin{center}
        {\LARGE\bfseries
        ClusIR: Towards Cluster-Guided All-in-One Image Restoration\par}
        \vspace{0.45em}
        {\Large\bfseries Supplementary Material\par}
    \end{center}
    \vspace{1.2em}
\end{@twocolumnfalse}
]

% Reset counters for the supplementary material.
\setcounter{section}{0}
\setcounter{subsection}{0}
\setcounter{figure}{0}
\setcounter{table}{0}
\setcounter{equation}{0}

% Optional S-prefix numbering. Remove these five lines if you
% prefer the default appendix numbering (A, B, ...).
\renewcommand{\thesection}{S\arabic{section}}
\renewcommand{\thesubsection}{\thesection.\arabic{subsection}}
\renewcommand{\thefigure}{S\arabic{figure}}
\renewcommand{\thetable}{S\arabic{table}}
\renewcommand{\theequation}{S\arabic{equation}}

\section{Experimental Settings}
% \subsection{Datasets} 
\noindent\textbf{One-by-One Degradations Setting.}
For the single task image restoration, we follow the standard evaluation protocols and construct three independent restoration tasks: denoising, deraining, and dehazing. 
For image denoising, we combine the BSD400~\cite{BSD400} and WED~\cite{WED} datasets as the training set, where BSD400 contains 400 images and WED provides 4{,}744 images. The images are corrupted with Gaussian noise at levels $\sigma\in\{15,25,50\}$. The denoising performance is evaluated on CBSD68~\cite{BSD68} and Urban100~\cite{Urban100}. 
For deraining, we adopt the Rain100L~\cite{Rain100L} dataset, which includes 200 image pairs for training and 100 pairs for testing. 
For dehazing, we use the SOTS~\cite{SOTS} dataset, consisting of 72{,}135 training images and 500 testing images. Each task is trained individually under this setting.

\noindent\textbf{Three Degradations Setting.}
For the All-in-One ``\textcolor{blue}{\bf{N}}+\textcolor{green}{\bf{H}}+\textcolor{red}{\bf{R}}'' configuration, we integrate the three degradation tasks—denoising, deraining, and dehazing—into a unified model. The training set is constructed by merging the BSD400+WED (denoising), Rain100L (deraining), and SOTS (dehazing) datasets. The model is trained for 150 epochs on this combined dataset and evaluated on the corresponding test sets for each restoration task.

\noindent\textbf{Five Degradations Setting.}
The Five-Degradation ``\textcolor{blue}{\textbf{N}}+\textcolor{green}{\textbf{H}}+\textcolor{red}{\bf{R}}+\textcolor{purple}{\textbf{B}}+\textcolor{pink}{\textbf{L}}'' setting builds upon the Three-Degradation setting, adding two additional tasks: deblurring and low-light enhancement.
For deblurring, we adopt the GoPro dataset~\cite{GoPro}, which consists of 2,103 training images and 1,111 testing images. For low-light enhancement, we use the LOL-v1 dataset~\cite{LOL}, containing 485 training images and 15 testing images. It is important to note that for the denoising task under the Five-Degradation setting, we report results using Gaussian noise with $\sigma=25$. The training takes 150 epochs.

\noindent\textbf{Composited Degradation Setting.}
For the composite degradation setting, we adopt the CDD11~\cite{OneRestore} dataset. CDD11 contains 1,183 training images, including:
(i) four single-degradation types: haze (H), low-light (L), rain (R), and snow (S);
(ii) five double-degradation types: low-light + haze (L+H), low-light + rain (L+R), low-light + snow (L+S), haze + rain (H+R), and haze + snow (H+S);
(iii) two triple-degradation types: low-light + haze + rain (L+H+R) and low-light + haze + snow (L+H+S). We train the model for 200 epochs while keeping all other settings unchanged.

\noindent\textbf{Real-world Degradation Setting.}
For the real-world degradation setting, we adopt the WeatherBench~\cite{weatherbench} dataset. It contains paired degraded–clean images under three representative weather conditions: \textit{rain}, \textit{snow}, and \textit{haze}. The dataset comprises 42{,}002 image pairs (41{,}402 for training and 600 for testing) with a resolution of $512\times512$, collected from diverse outdoor scenes under both daytime and nighttime conditions. Compared with synthetic benchmarks, WeatherBench provides more realistic and naturally coupled degradation distributions, reducing domain gaps and enabling more reliable evaluation of model robustness and generalization. 
We train the model for 250 epochs while keeping all other settings unchanged.

% For the real-world degradation setting, we adopt the WeatherBench~\cite{weatherbench} dataset. WeatherBench exhibits naturally coupled and unknown degradation patterns, making it more representative of real-world scenarios. Therefore, it provides a challenging benchmark to evaluate the robustness and generalization capability of restoration methods under practical conditions.

\begin{figure*}[t]
    \centering
    \includegraphics[width=1\linewidth]{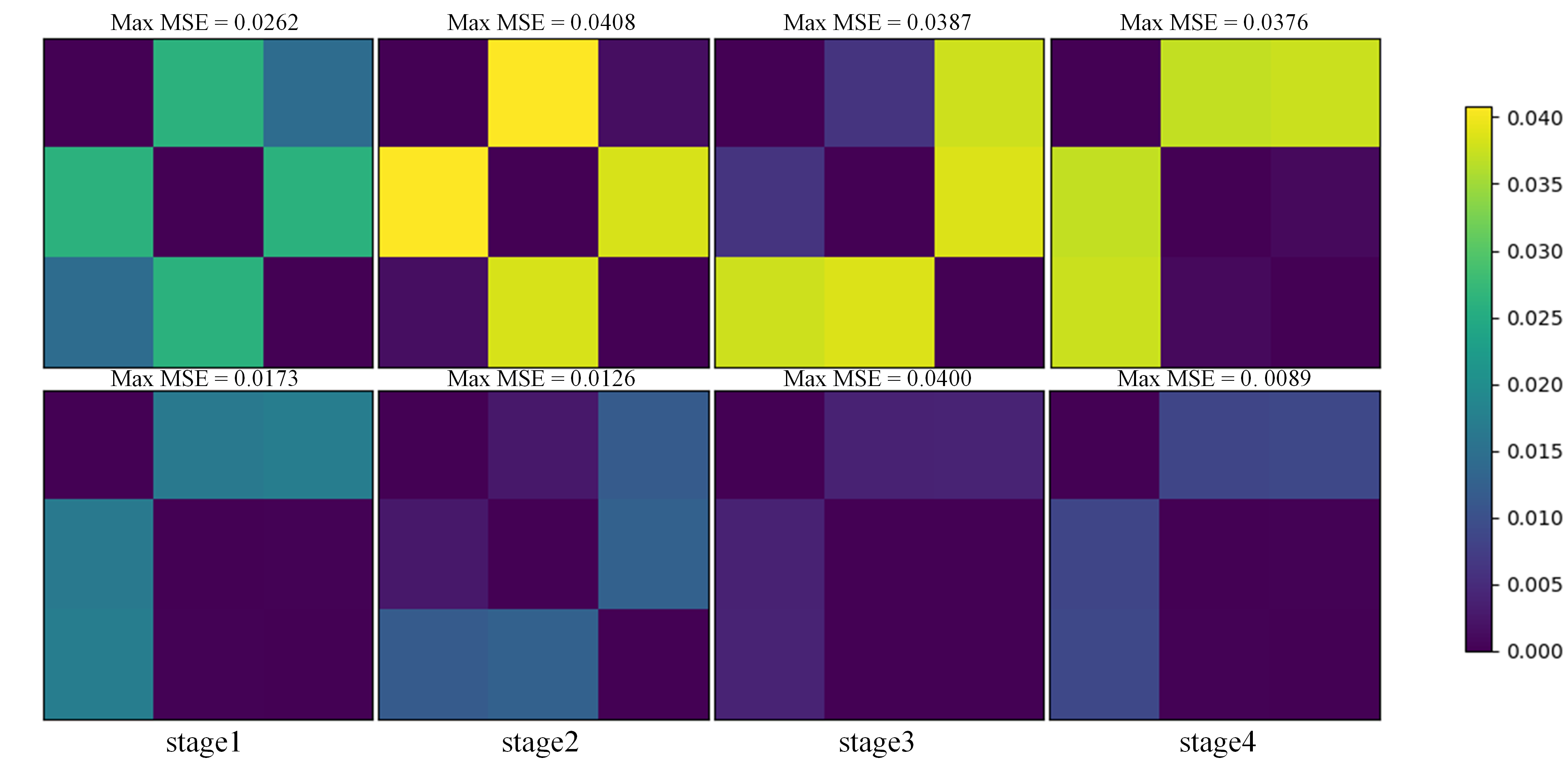}
    \caption{Pairwise prototype MSE across stages under orthogonal (top) and random (bottom) initialization. Orthogonal initialization yields consistently larger inter-prototype distances, indicating better degradations separability.}
    \label{mse_ana}
\end{figure*}

\section{All-in-One Restoration Result}
\begin{table}[t]
\centering
\small
\renewcommand\arraystretch{1}
\setlength{\tabcolsep}{2pt}
\caption{LPIPS comparison under AiOIR-3 setting. Lower is better.}
\begin{tabular}{l|ccc|c|c|c}
\hline\hline
\textbf{Method} 
& \multicolumn{3}{c|}{\textbf{Denoise}} 
& \begin{tabular}{c}
\textbf{Dehaze} 
\end{tabular}
& \begin{tabular}{c}
\textbf{Derain} 
\end{tabular}
& \textbf{Avg.} \\
\cline{2-4}
& $\sigma{=}15$ & $\sigma{=}25$ & $\sigma{=}50$ 
& & & \\
\hline
PromptIR \cite{PromptIR}
& 0.1571 & 0.2216 & 0.3174 & 0.0456 & 0.0329 & 0.1549 \\

MoCE-IR \cite{moceir}
& 0.1412 & 0.1996 & 0.2927 & 0.0338 & 0.0270 & 0.1380 \\

\textbf{ClusIR} 
& \textbf{0.1401} & \textbf{0.1932} & \textbf{0.2799} 
& \textbf{0.0286} & \textbf{0.0240} & \textbf{0.1332} \\
\hline\hline
\end{tabular}
\label{tab:lpips_three_task}
\end{table}
% \begin{table}[t]
% % \vspace{-3mm}
% \centering
% \renewcommand\arraystretch{1}
% \setlength{\tabcolsep}{3pt}
% \caption{LPIPS comparison under AiOIR-3 setting. Lower is better.}
% \resizebox{\linewidth}{!}{
% \begin{tabular}{l|ccc|c|c|c}
% \hline\hline
% \textbf{Method} 
% & \multicolumn{3}{c|}{\textbf{Denoise (CBSD68)}} 
% & \textbf{Dehaze (SOTS)} 
% & \textbf{Derain (Rain100L)} 
% & \textbf{Avg.} \\
% \cline{2-4}
% & $\sigma{=}15$ & $\sigma{=}25$ & $\sigma{=}50$ 
% & & & \\
% \hline
% PromptIR \cite{PromptIR}
% & 0.1571 & 0.2216 & 0.3174 & 0.0456 & 0.0329 & 0.1549 \\

% MoCE-IR \cite{moceir}
% & 0.1412 & 0.1996 & 0.2927 & 0.0338 & 0.0270 & 0.1380 \\

% \textbf{ClusIR} 
% & \textbf{0.1401} & \textbf{0.1932} & \textbf{0.2799} 
% & \textbf{0.0286} & \textbf{0.0240} & \textbf{0.1332} \\
% \hline\hline
% \end{tabular}
% }
% % \vspace{-3mm}
% \label{tab:lpips_three_task}
% \end{table}

To comprehensively evaluate the perceptual quality in the all-in-one image restoration setting, we compare ClusIR with representative methods using the LPIPS metric under the three degradation setting.

Tab.~\ref{tab:lpips_three_task} reports the quantitative comparison in terms of LPIPS under the three degradation setting. ClusIR consistently achieves the best performance (0.1332) across all degradations, reducing LPIPS by 0.0217 and 0.0048 compared to PromptIR (0.1549) and MoCE-IR (0.1380), respectively. These improvements stem from the proposed PCGRM and DAFMM, which facilitate cluster-guided degradation modeling and adaptive frequency modulation. As a result, ClusIR achieves higher perceptual quality under complex degradation scenarios.

% We extend the previous AiOIR settings by constructing various composite degradation scenarios, resulting in eleven distinct restoration settings in total.

% \noindent\textbf{Results Comparisons on Composited Degradations:}
% Tab.~\ref{cdd} presents a comparison of ClusIR with state-of-the-art methods on the CDD11 dataset under Composited Degradation setting. ClusIR outperforms existing methods in terms of both PSNR and SSIM across all degradation combinations. Specifically, ClusIR achieves the highest performance, with an average PSNR of 27.33 dB and SSIM of 0.878, surpassing AirNet~\cite{AirNet}, PromptIR~\cite{PromptIR}, WGWSNet~\cite{WGWSNet}, and WeatherDiff~\cite{WeatherDiff} by gains of +3.58 dB / +0.075, +1.43 dB / +0.028, +0.37 dB / +0.015, and +4.84 dB / +0.079, respectively.
% Notably, for the CDD11-Double and CDD11-Triple settings, ClusIR consistently delivers superior performance, highlighting its robustness in handling multiple types of image degradation. 
% Overall, these results demonstrate ClusIR's effectiveness in all-in-one image restoration, showcasing its ability to simultaneously address multiple degradation tasks with a single unified model. These improvements can be attributed to our proposed PCGRM, which not only enables effective degradation discrimination but also activates the corresponding expert groups, facilitating the integration of specialized knowledge to handle diverse degradation types efficiently.

\section{Ablation Studies}
We conduct several ablation experiments to demonstrate the effectiveness of our proposed ClusIR. We report the results under the Three Degradations Setting.

\begin{table}[t]
\caption{Details of our base, small, and tiny variants of ClusIR. 
FLOPs are computed on an input of size $128\times128$ using a NVIDIA 3060 GPU.}
\centering
% \vspace{0.3em}
\renewcommand\arraystretch{1.18}
\resizebox{\linewidth}{!}{
\begin{tabular}{l|c|c|c}
\hline\hline
\textbf{Architecture} & \textbf{ClusIR-Base} & \textbf{ClusIR-S} & \textbf{ClusIR-T} \\
\hline
% The Number of Encoder Blocks & [4,6,6,8] & [4,6,6,8] & [4,6,6,8] \\
The Input Embedding Dimension & 48 & 32 & 16 \\
\hline
Params. & 54.66M & 25.03M & 6.89M \\
FLOPs  & 28.85G & 13.42G & 3.79G \\
\hline\hline
\end{tabular}}
% \vspace{-0.4em}
\label{scale}
% \vspace{-1.5em}
\end{table}

\begin{table}[t]
\caption{Performance of different ClusIR (T/S/Base) on three degradation tasks. 
PSNR and SSIM are reported on full RGB images.}
\centering
\renewcommand\arraystretch{1}
\setlength{\tabcolsep}{1pt}
\resizebox{\linewidth}{!}{
\begin{tabular}{l|ccc|c|c|c}
\hline\hline
\textbf{Architecture} 
& \multicolumn{3}{c|}{\textbf{Denoise}} 
& \textbf{Derain} 
& \textbf{Dehaze} 
& \textbf{Avg.} \\
\cline{2-4}
& $\sigma{=}15$ & $\sigma{=}25$ & $\sigma{=}50$ 
& & & \\
\hline
\textbf{ClusIR-T} 
& 33.70/0.933 & 31.06/0.890 & 27.77/0.806 
& 35.44/0.970 & 31.92/0.981 & 31.98/0.916 \\
\textbf{ClusIR-S} 
& 34.07/0.937 & 31.41/0.896 & 29.11/0.813 
& 37.38/0.980 & 33.06/0.981 & 32.21/0.921 \\
\textbf{ClusIR-Base} 
& 34.10/0.938 & 31.45/0.897 & 28.19/0.814 
& 38.71/0.984 & 32.85/0.983 & 33.06/0.923 \\
\hline\hline
\end{tabular}
}
% \vspace{-1.5em}
\label{model_per}
\end{table}

\begin{table}[t]
\centering
\caption{Results on $128$ patches by the RTX 3060 GPU.}
% \vspace{-0.6em}
\setlength{\tabcolsep}{2.7pt}
\renewcommand{\arraystretch}{1}
\begin{tabular}{l|c|c|c|c}
\hline\hline
\textbf{Method} 
& \begin{tabular}{c}
\textbf{Params} \\
\textbf{(M)}
\end{tabular}
& \begin{tabular}{c}
\textbf{FLOPs} \\
\textbf{(G)}
\end{tabular}
& \begin{tabular}{c}\textbf{Inference} \\ \textbf{Time (ms)}\end{tabular} 
& \textbf{Avg.} \\
\hline
AirNet \cite{AirNet}            
& 8.93 & 73.42 & 89.488$\pm$0.899 & 31.20/0.910 \\
PromptIR \cite{PromptIR}            
& 35.59 & 43.18 & 75.723$\pm$6.912 & 32.06/0.913 \\
% MoCE-IR  \cite{moceir}       
% & 25.35 & 25.81 & 68.579$\pm$1.712 & 32.73/0.917 \\
\hline
ClusIR-T              
& 6.89 & 3.34 & 68.851$\pm$3.130 & 31.98/0.916 \\
ClusIR-S              
& 25.03 & 12.53 & 107.698$\pm$8.941 & 32.21/0.921 \\
ClusIR-Base              
& 54.66 & 28.85 & 118.311$\pm$2.001 & 33.06/0.923 \\
\hline\hline
\end{tabular}
\label{tab:efficiency}
\end{table}
\noindent\textbf{Model Scaling.} We propose two variants of ClusIR with different scales, namely Small (ClusIR-S) and Tiny (ClusIR-T), as shown in Tab.~\ref{scale}.

\noindent\textbf{Scaling Analysis.} 
As illustrated in Tab.~\ref{model_per}, ClusIR-T achieves a significant reduction in computational cost compared to the base model, with only 6.89M parameters and 3.79G FLOPs. Despite this, it maintains competitive performance, achieving 31.98dB/0.916. ClusIR-S strikes a balance between model size and computational efficiency with 25.03M parameters and 13.42G FLOPs. Notably, ClusIR-T surpasses AirNet~\cite{AirNet} by +0.78dB/0.006 while ClusIR has fewer parameters.

\noindent\textbf{Efficiency Comparison.}Tab.~\ref{tab:efficiency} shows that ClusIR-S outperforms PromptIR\cite{PromptIR} on three degradations setting with substantially fewer parameters and FLOPs. ClusIR-T further reduces computational cost and inference time compared to AirNet\cite{AirNet}, while achieving superior performance. These results demonstrate that ClusIR achieves superior restoration performance while reducing computational cost.

% These results suggest that the performance gains are primarily attributed to the proposed PCGRM and DAFMM, rather than increased model capacity.

% Moreover, ClusIR-Base surpasses MoCE-IR\cite{moceir} under comparable computational budgets. 
% Tab.~\ref{tab:efficiency} shows that ClusIR-S outperforms PromptIR on AiOIR-3 with substantially fewer parameters/FLOPs; ClusIR-T significantly reduces complexity and inference time vs.\ AirNet while achieving better performance; and ClusIR-Base outperforms MoCE-IR at comparable computational cost. Tab.~\ref{tab:efficiency} indicates that the gains are primarily driven by the proposed PCGRM and DAFMM, rather than merely by increased model capacity.

\begin{table}[t]
\caption{Impact of Orthogonal Initialization on Model Performance.}
\centering
\setlength{\tabcolsep}{1pt}
\renewcommand\arraystretch{1}
\resizebox{\linewidth}{!}{
\begin{tabular}{l|ccc|c|c|c}
\hline\hline
\textbf{Initialization} 
& \multicolumn{3}{c|}{\textbf{Denoise}} 
& \textbf{Derain} 
& \textbf{Dehaze} 
& \textbf{Avg.} \\
\cline{2-4}
& $\sigma{=}15$ & $\sigma{=}25$ & $\sigma{=}50$ 
& & & \\
\hline
\textbf{Orthogonal}  
& 34.10/0.938 & 31.45/0.897 & 28.19/0.814 
& 38.71/0.984 & 32.85/0.983 & 33.06/0.923 \\
\textbf{Random} 
& 34.12/0.937 & 31.45/0.897 & 28.17/0.813 
& 38.28/0.983 & 32.16/0.982 & 32.84/0.922 \\
\hline\hline
\end{tabular}
}
\label{ore_res}
\vspace{-2em}
\end{table}

\noindent\textbf{Orthogonal Initialization.} To evaluate the effect of orthogonal initialization, we replace it with uniformly random prototype initialization for comparison.

As shown in Tab.~\ref{ore_res}, orthogonal initialization consistently improves performance across all tasks, achieving an average PSNR/SSIM of 33.06/0.923 and clearly surpassing the randomly initialized counterpart.
This advantage stems from the strengthened degradation discrimination induced by orthogonal cluster prototypes. As visualized in Fig.~\ref{mse_ana}, the maximum inter-cluster MSE under orthogonal initialization reaches 0.0262/0.0408 in Stage~1 and Stage~2, respectively, whereas the non-orthogonal counterpart yields only 0.0173/0.0126. The orthogonal initialization produces significantly larger inter-prototype distances, where degradation cues are most distinct. In deeper stages, the prototypes gradually exhibit the emergence of semantic structure while still preserving meaningful separation among degradation types, indicating that orthogonality provides a stable prior throughout the hierarchy. Overall, the proposed orthogonal initialization enhances prototype separability, reduces semantic overlap among clusters, and leads to more discriminative representations.

% We further examine the pairwise distances among cluster prototypes by computing the MSE matrices across all PCGRM modules. With orthogonal initialization, the off-diagonal entries remain consistently larger (e.g., 0.0262–0.0408 in early stages), indicating strong inter-cluster separation throughout the hierarchy. In contrast, randomly initialized prototypes exhibit much smaller and uneven distances (typically 0.002–0.017), revealing significant overlap among clusters.
% This contrast confirms that orthogonal initialization enforces well-spread prototype directions in the embedding space, stabilizing the semantic boundaries between degradation clusters.
% Such improved separability directly benefits the routing mechanism of PCGRM, enabling more reliable expert assignment and ultimately leading to the performance gains observed in Table~\ref{meodel_per}.

\begin{table}[t]
    \centering
     % \vspace{-4em}
    \caption{Key distinctions between DAFMM and prior frequency-based modules.}
    \resizebox{\linewidth}{!}{%
    \begin{tabular}{lcc}
        \toprule\toprule
         & \textbf{Prior freq.-based mod./decomp.} & \textbf{DAFMM (Ours)} \\
        \midrule
        Conditioning  & Implicit / task-agnostic  & Cluster-level degradation semantics \\
        Freq.\ transform & Fixed (e.g., FFT/DCT/Wavelet) & Fixed transform + learnable filtering \\
        Modulation target & Often implicit / coarse & Adaptive low/high-frequency modulation \\
        Link to routing & Typically separate from routing & Semantic-to-frequency bridge via PCGRM \\
        \bottomrule\bottomrule
    \end{tabular}}
    \vspace{-1em}
    \label{tab:dafmm_distinction}
\end{table}

\subsection{Comparison between DAFMM and Prior Methods}
As summarized in Tab.~\ref{tab:dafmm_distinction}, DAFMM differs from prior frequency-based methods by explicitly conditioning frequency modulation on cluster-level degradation semantics rather than implicit, task-agnostic learning. Through cluster-guided routing, these semantics are translated into frequency-domain modulation, tightly coupling frequency refinement with routing. Beyond fixed frequency transforms, DAFMM combines wavelet decomposition with learnable frequency filtering to adaptively refine low- and high-frequency components, functioning as a semantic-to-frequency bridge.

\begin{figure}[t]
    \centering
    % \vspace{-1em}
    \includegraphics[width=1\linewidth]{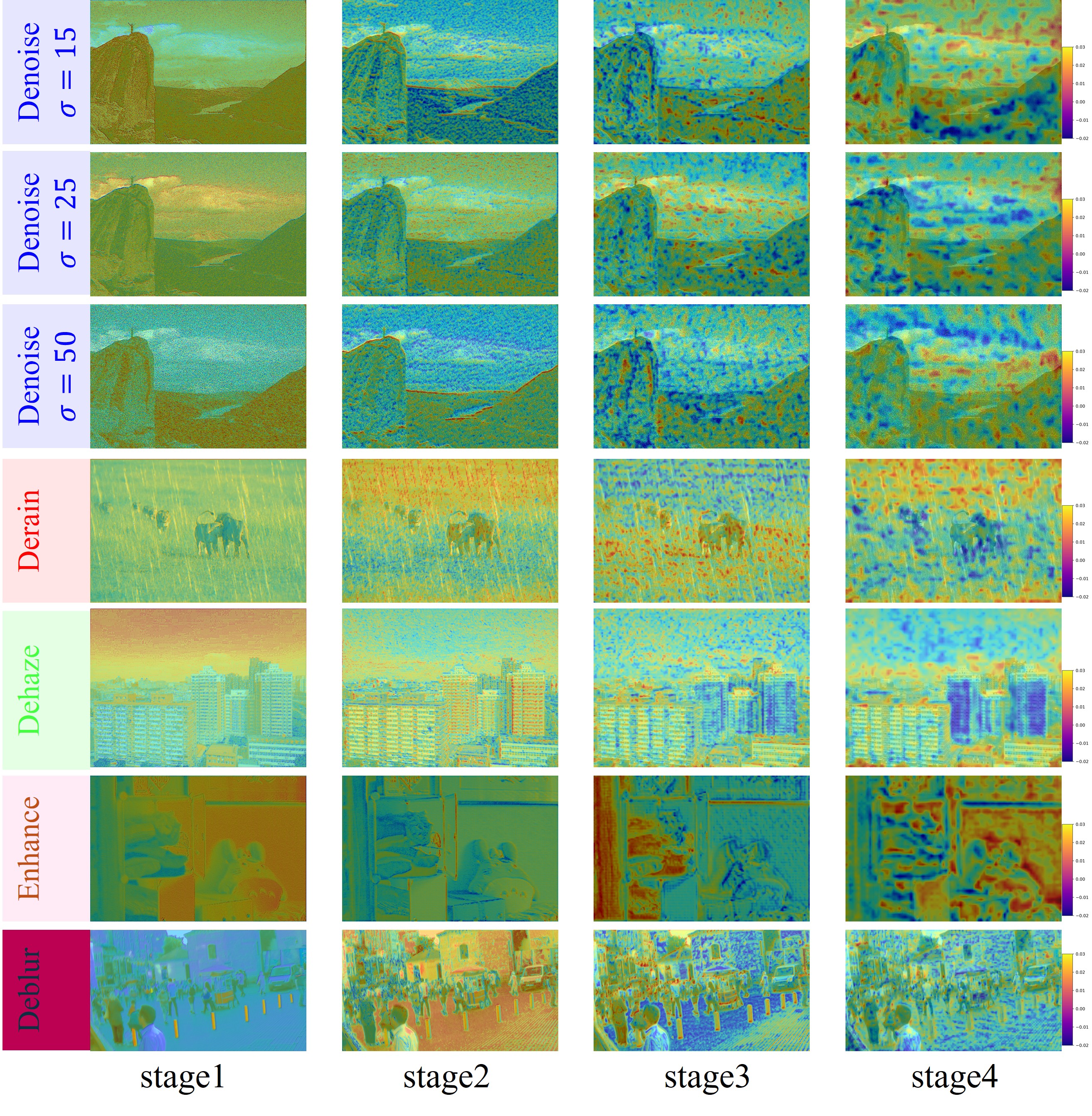}
       % \vspace{-2em}
    \caption{Visual degradation of PCGRM-MoE under the (\textcolor{blue}{\textbf{N}}+\textcolor{green}{\textbf{H}}+\textcolor{red}{\bf{R}}+\textcolor{purple}{\textbf{B}}+\textcolor{pink}{\textbf{L}}) setting. Higher values mean stronger correlations.}
    \label{fig:5task_heatmap}
    \vspace{-2.5em}
\end{figure}

\section{Visualization}
% We provide more visual comparisons for the One-by- One Degradations setting, Three Degradations setting and a more complex Five Degradations settings.
We provide more visual comparisons under four evaluation settings: the one-by-one degradation setting, the three-degradation setting, the five-degradation setting, and a more challenging composited degradation setting.

\subsection{Visual Comparison under One Degradations}
\begin{figure*}[!h]
    \centering
    \includegraphics[width=\linewidth]{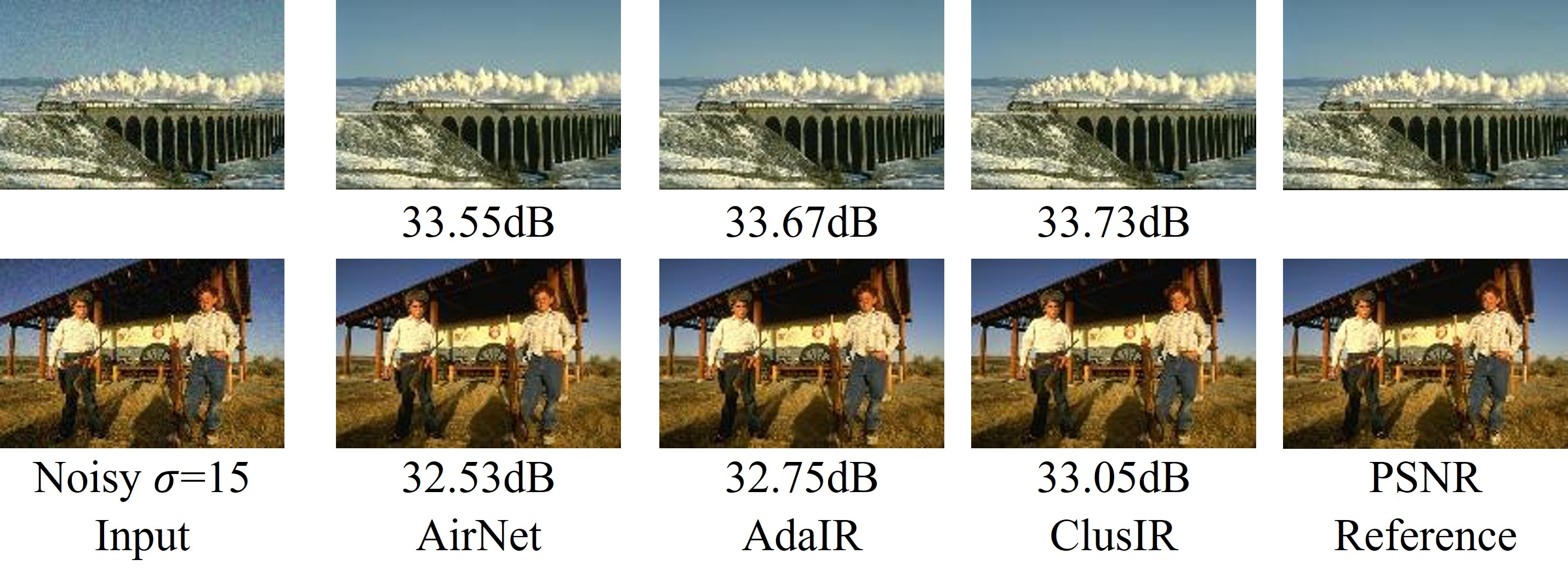}
     \vspace{-2.5em}
    \caption{Denoising ($\sigma=15$) visual comparisons of ClusIR with state-of-the-art All-in-One methods under One-by-One setting.}
    \label{fig:1tasks_denoise15}
\end{figure*}

\begin{figure*}[!h]
    \centering
    \includegraphics[width=\linewidth]{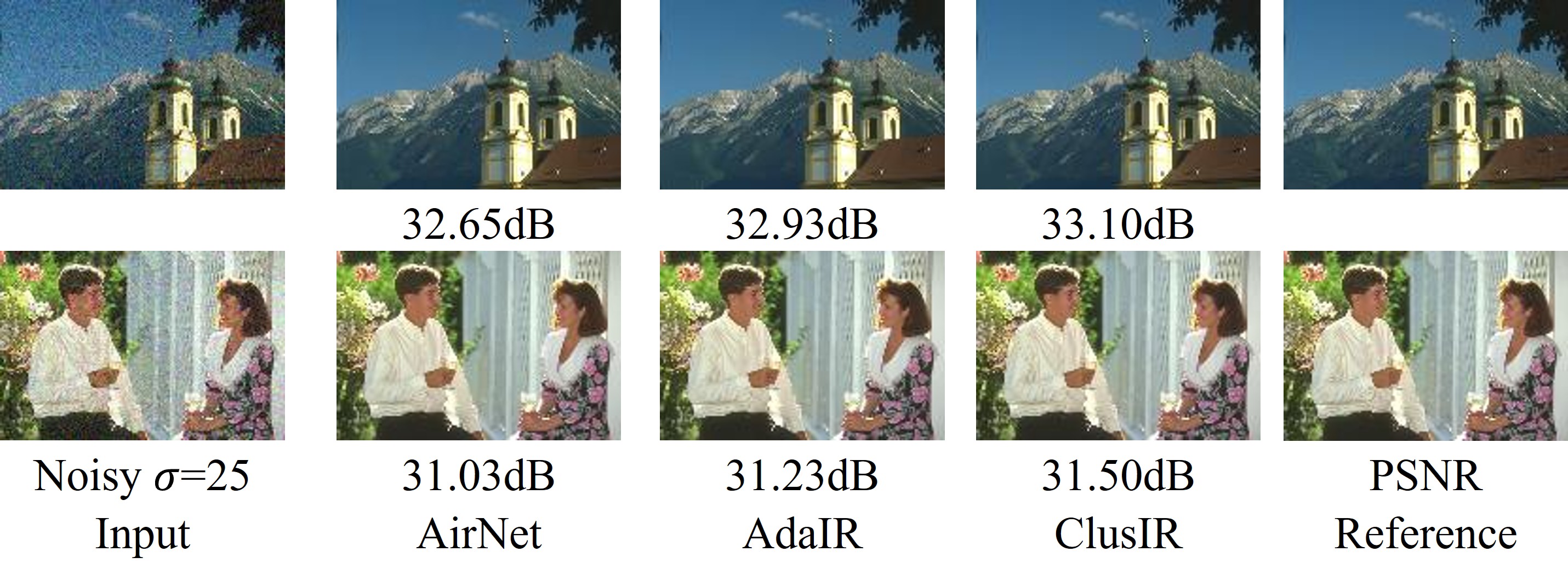}
    \vspace{-2.5em}
    \caption{Denoising ($\sigma=25$) visual comparisons of ClusIR with state-of-the-art All-in-One methods under One-by-One setting.}
    \vspace{-1em}
    \label{fig:1tasks_denoise25}
\end{figure*}

\begin{figure*}[!h]
    \centering
    \includegraphics[width=\linewidth]{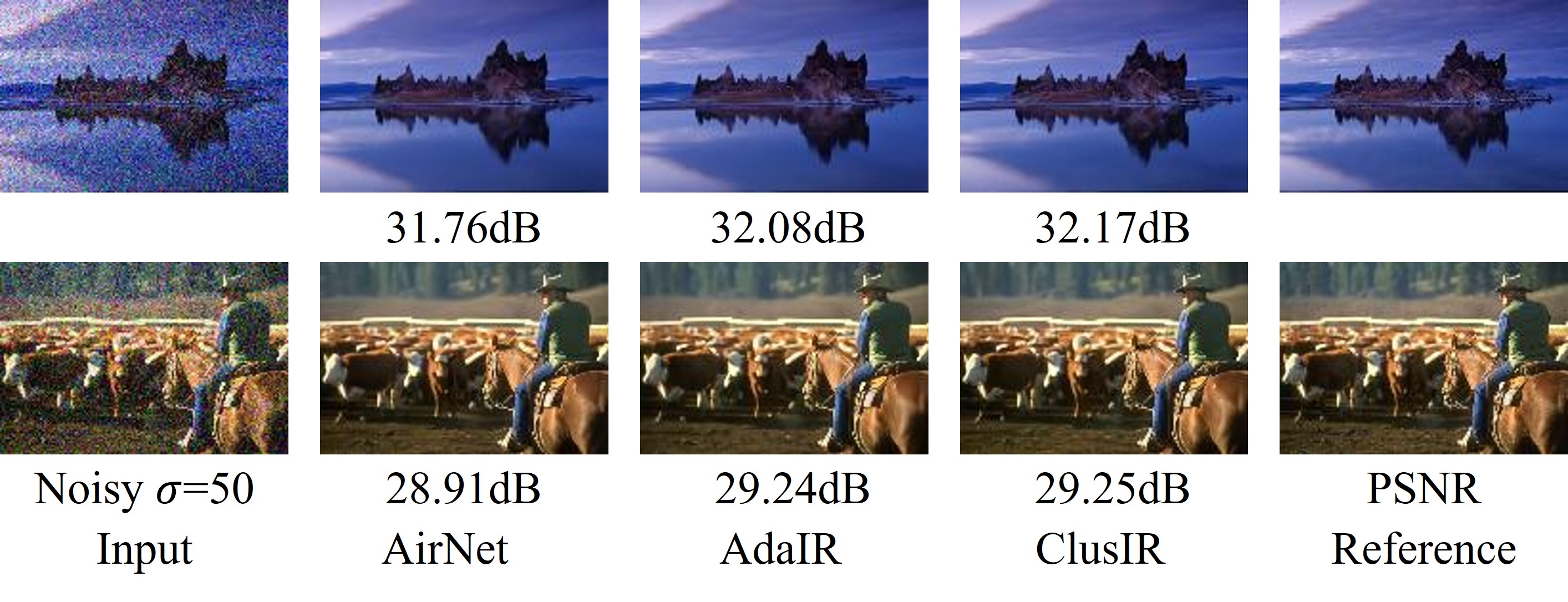}
        \vspace{-2.5em}
    \caption{Denoising ($\sigma=50$) visual comparisons of ClusIR with state-of-the-art All-in-One methods under One-by-One setting.}
    \label{fig:1tasks_denoise50}
\end{figure*}

\begin{figure*}[!h]
    \centering
    \includegraphics[width=1\linewidth]{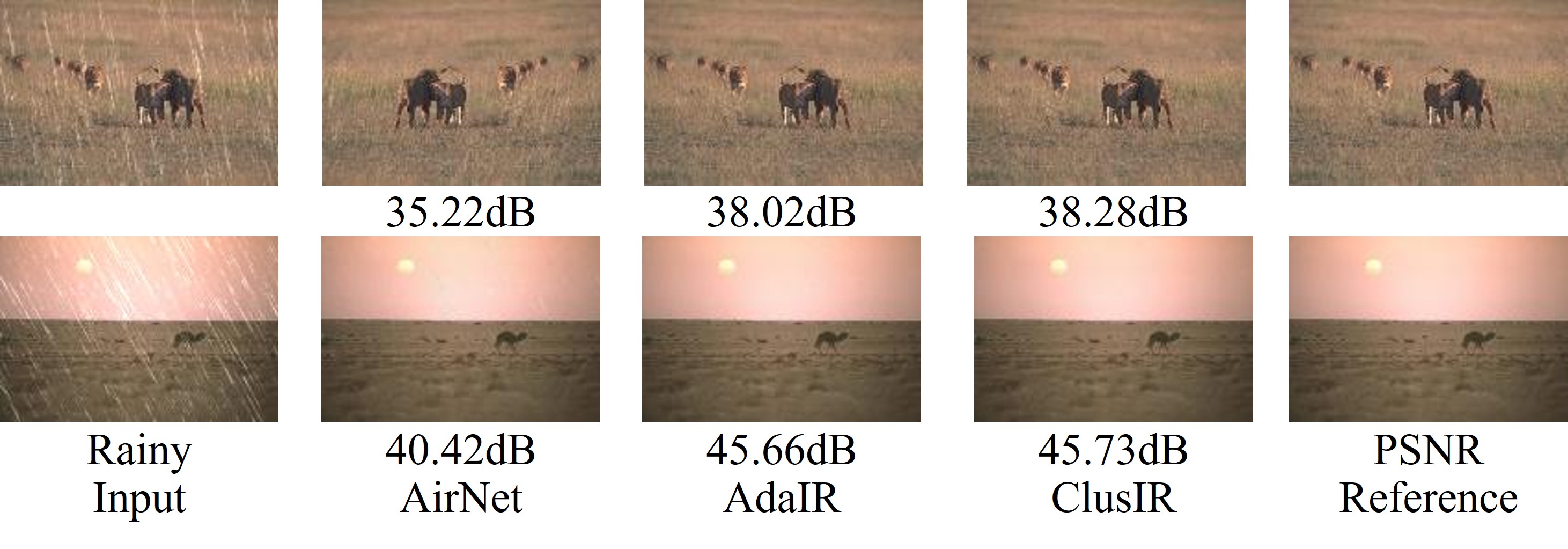}
    \vspace{-3em}
    \caption{Deraining visual comparisons of ClusIR with state-of-the-art All-in-One methods under One-by-One setting.}
    \label{fig:1tasks_derain}
     \vspace{-1em}
\end{figure*}

\begin{figure*}[!h]
    \centering
    \includegraphics[width=1\linewidth]{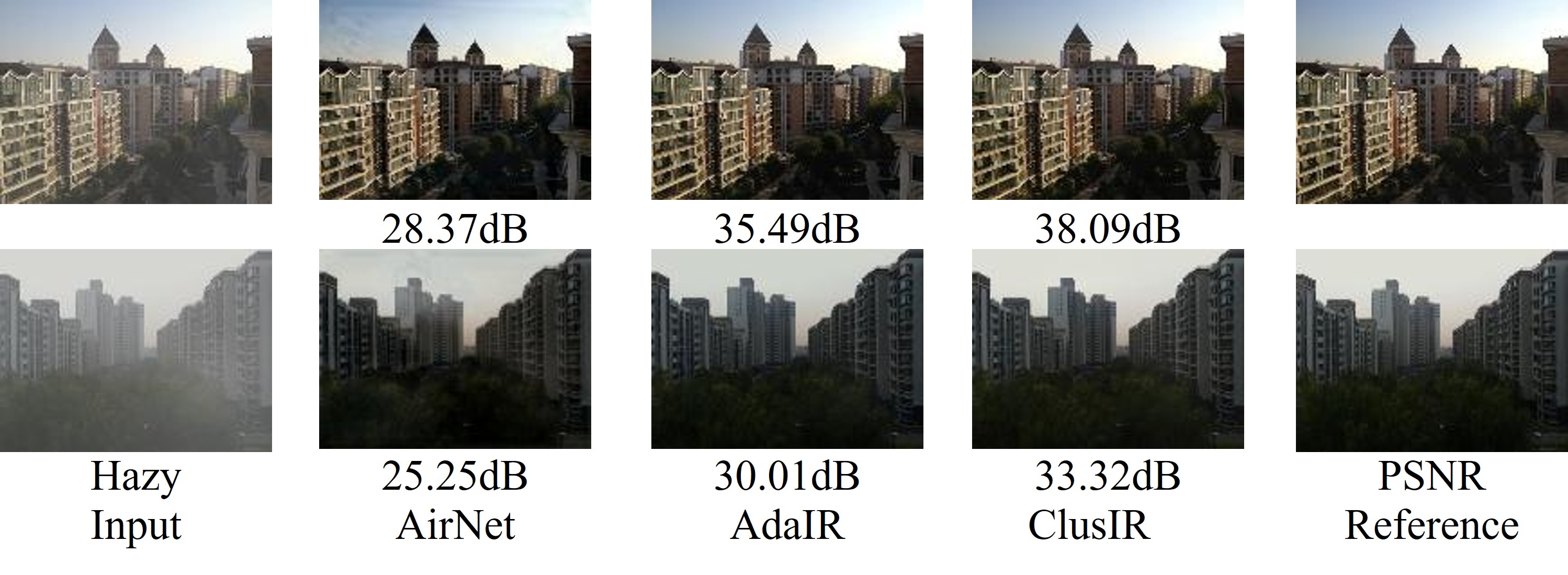}
    \vspace{-3em}
    \caption{Dehazing visual comparisons of ClusIR with state-of-the-art All-in-One methods under One-by-One setting.}
    \label{fig:1tasks_dehaze}
     \vspace{-1em}
\end{figure*}

As shown in Fig.~\ref{fig:1tasks_denoise15}, \ref{fig:1tasks_denoise25}, \ref{fig:1tasks_denoise50}, \ref{fig:1tasks_derain} and \ref{fig:1tasks_dehaze}, we compare our ClusIR against recent state-of-the-art methods, including AirNet~\cite{AirNet} (CVPR'22) and AdaIR~\cite{DBLP:conf/iclr/0001ZKKSK25} (ICLR’25).

Across all single degradation, ClusIR produces cleaner structures, sharper textures, and fewer residual artifacts, demonstrating that its cluster-guided modeling remains highly effective even when the degradation type is fixed.

\subsection{Visual Comparison under Three Degradations}
\begin{figure*}[!h]
    \centering
    \includegraphics[width=1\linewidth]{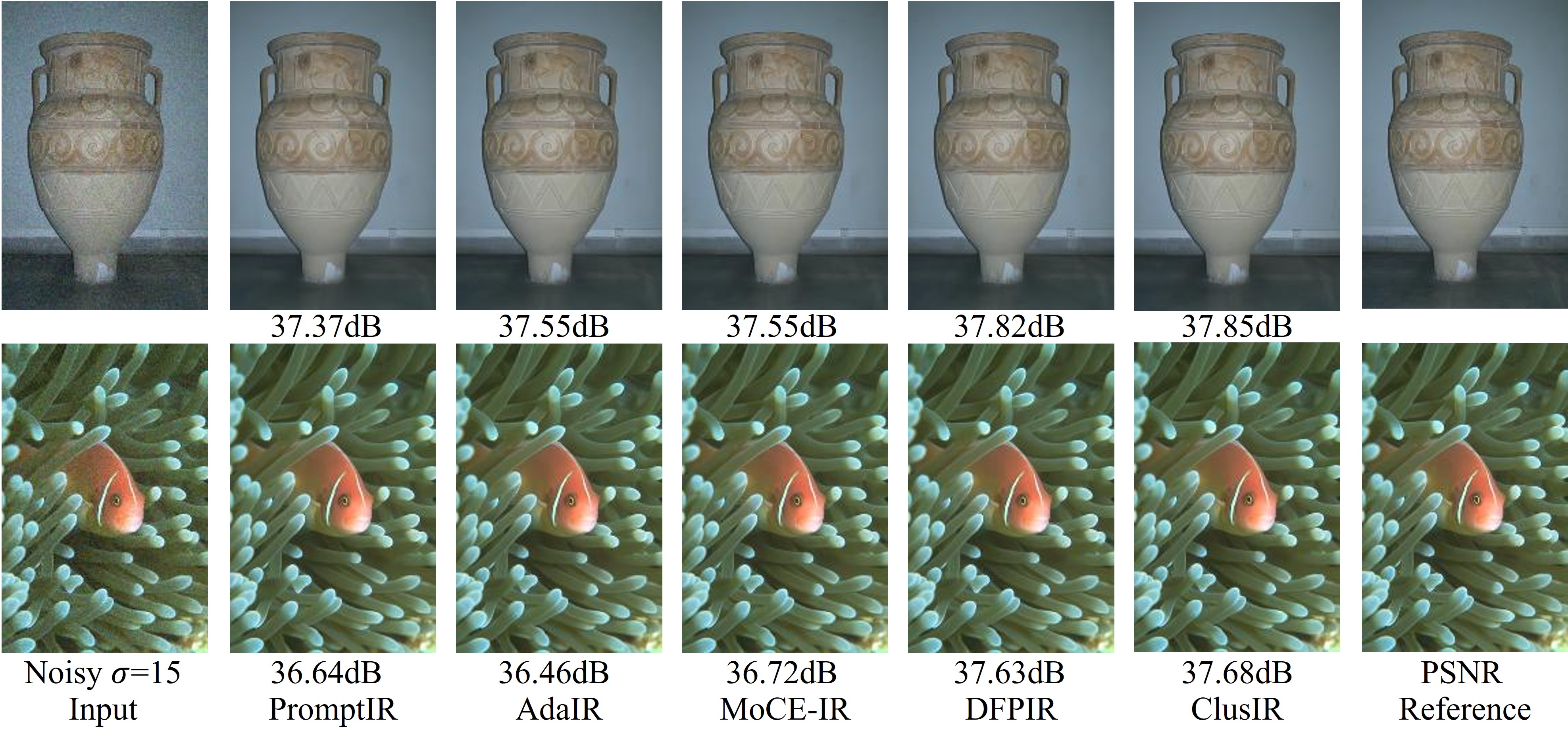}
    \vspace{-3em}
    \caption{Denoising ($\sigma = 15$) visual comparisons of ClusIR with state-of-the-art All-in-One methods under ``\textcolor{blue}{\bf{N}}+\textcolor{green}{\bf{H}}+\textcolor{red}{\bf{R}}'' setting.}
    \label{fig:3tasks_denoise15}
    \vspace{-1em}
\end{figure*}

\begin{figure*}[!h]
    \centering
    \includegraphics[width=1\linewidth]{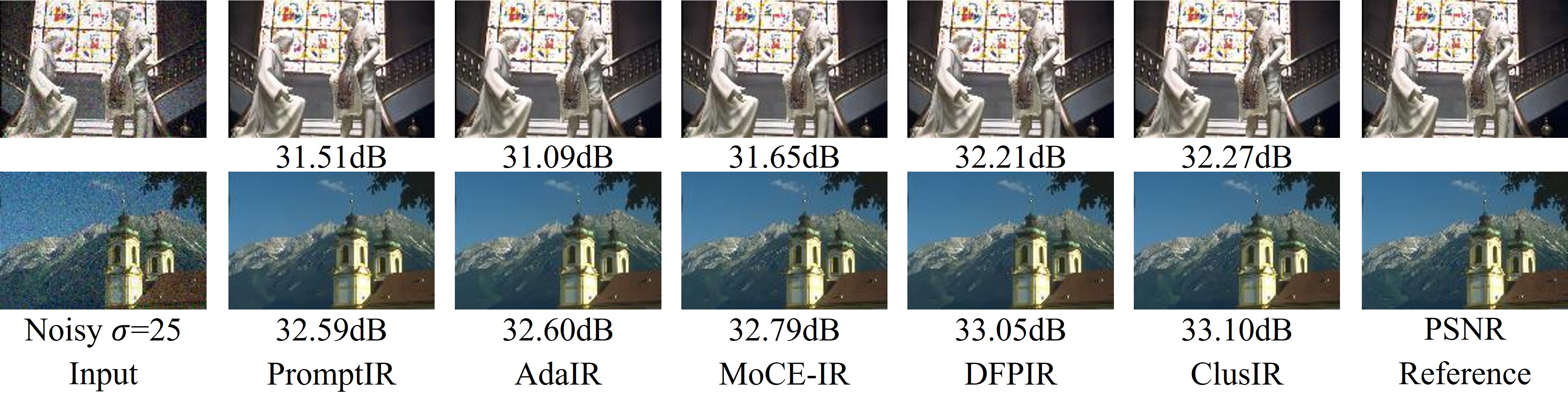}
    \vspace{-2.5em}
    \caption{Denoising ($\sigma = 25$) visual comparisons of ClusIR with state-of-the-art All-in-One methods under ``\textcolor{blue}{\bf{N}}+\textcolor{green}{\bf{H}}+\textcolor{red}{\bf{R}}'' setting.}
    \label{fig:3tasks_denoise25}
\end{figure*}

\begin{figure*}[!h]
    \centering
    \includegraphics[width=1\linewidth]{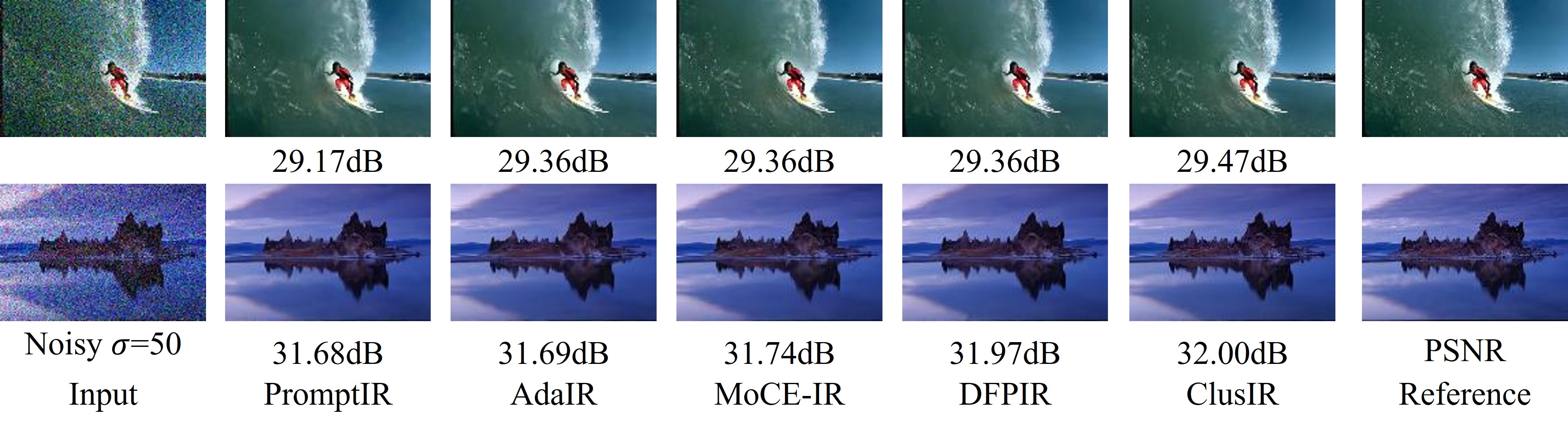}
    \vspace{-2.5em}
    \caption{Denoising ($\sigma = 50$) visual comparisons of ClusIR with state-of-the-art All-in-One methods under ``\textcolor{blue}{\bf{N}}+\textcolor{green}{\bf{H}}+\textcolor{red}{\bf{R}}'' setting.}
    \label{fig:3tasks_denoise50}
\end{figure*}

\begin{figure*}[!h]
    \centering
    \includegraphics[width=1\linewidth]{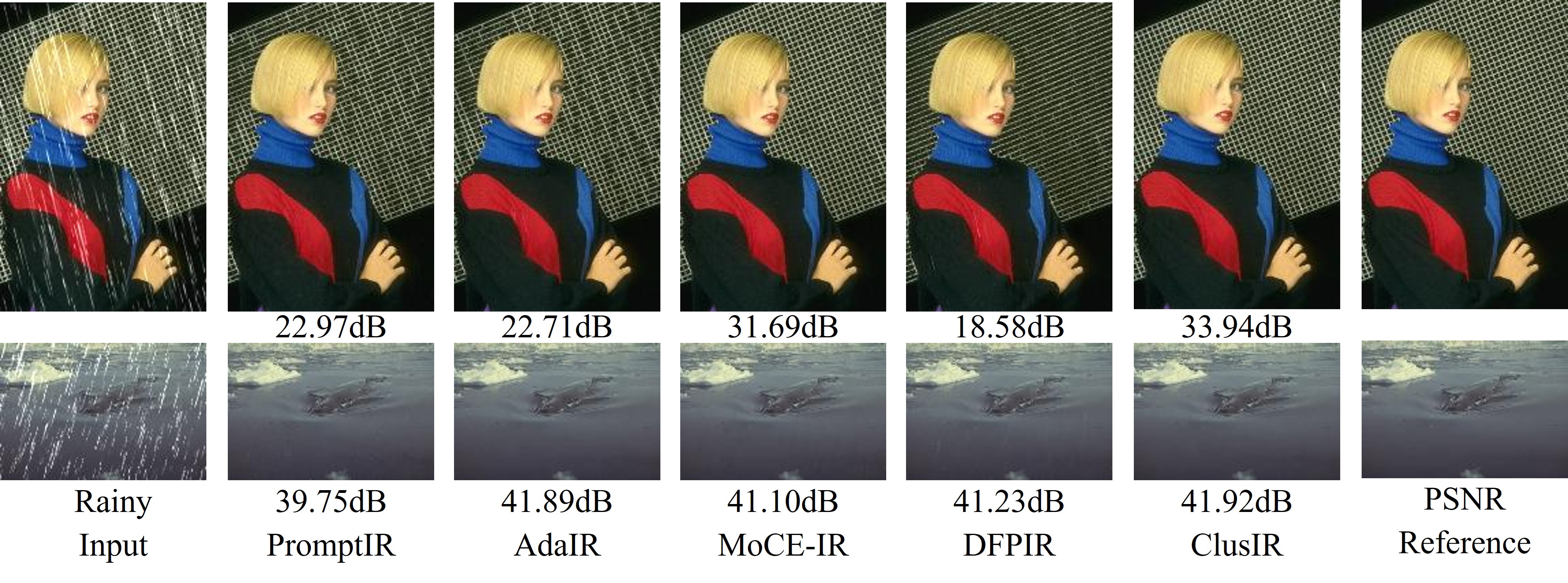}
    \vspace{-3em}
    \caption{Deraining visual comparisons of ClusIR with state-of-the-art All-in-One methods under ``\textcolor{blue}{\bf{N}}+\textcolor{green}{\bf{H}}+\textcolor{red}{\bf{R}}'' setting.}
    \label{fig:3tasks_derain}
\end{figure*}

\begin{figure*}[!h]
    \centering
    \includegraphics[width=1\linewidth]{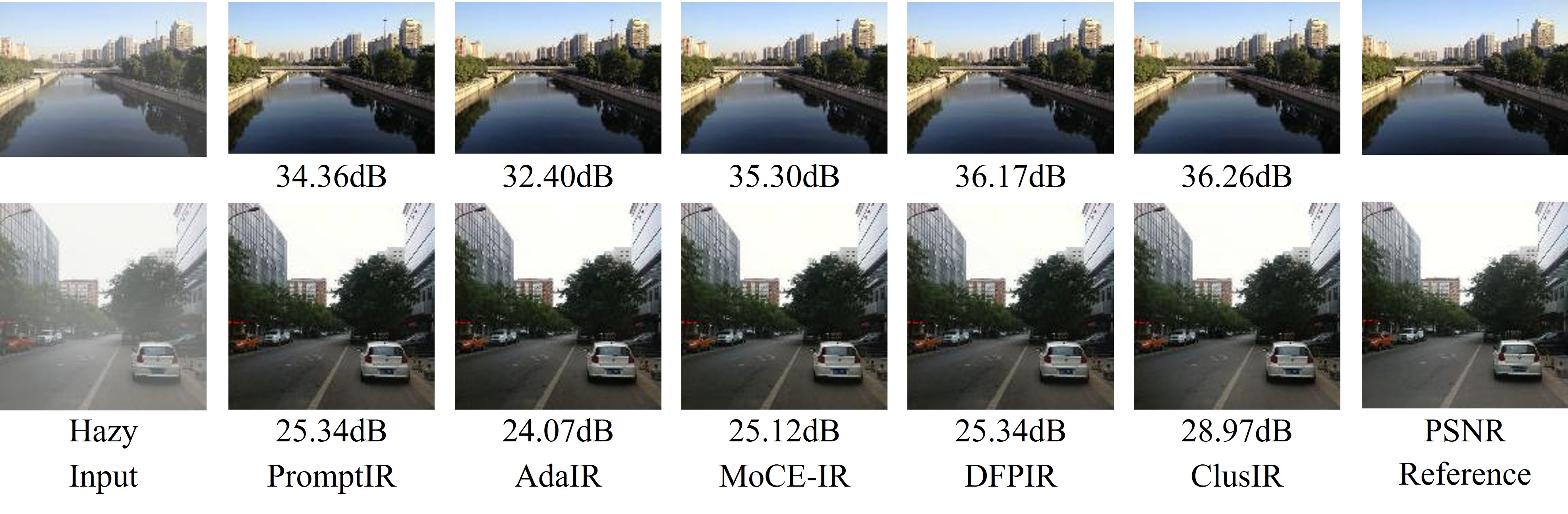}
    \vspace{-3em}
    \caption{Dehazing visual comparisons of ClusIR with state-of-the-art All-in-One methods under ``\textcolor{blue}{\bf{N}}+\textcolor{green}{\bf{H}}+\textcolor{red}{\bf{R}}'' setting.}
    \label{fig:3tasks_dehaze}
     \vspace{-1em}
\end{figure*}

As shown in Fig.~\ref{fig:3tasks_denoise15}, \ref{fig:3tasks_denoise25}, \ref{fig:3tasks_denoise50}, \ref{fig:3tasks_derain}, and \ref{fig:3tasks_dehaze}, we compare our ClusIR against recent state-of-the-art methods, including AdaIR~\cite{DBLP:conf/iclr/0001ZKKSK25} (ICLR’25), DFPIR~\cite{DBLP:conf/cvpr/TianLLLR25} (CVPR’25), and MoCE-IR~\cite{DBLP:conf/cvpr/ZamfirWMTP0T25} (CVPR’25).

Across all degradation types, ClusIR consistently preserves higher fidelity while maintaining fine textures and structural details. The zoomed-in regions further highlight ClusIR’s superiority in recovering sharp edges and realistic patterns compared to other approaches. These visual results are consistent with the quantitative improvements reported earlier.

\subsection{Visual Comparison under Five Degradations}
As some methods do not provide visual results under the five-degradation setting (\textcolor{blue}{\textbf{N}}+\textcolor{green}{\textbf{H}}+\textcolor{red}{\bf{R}}+\textcolor{purple}{\textbf{B}}+\textcolor{pink}{\textbf{L}}), we select InstructIR~\cite{InstructIR} (ECCV’ 24) and MoCE-IR~\cite{DBLP:conf/cvpr/ZamfirWMTP0T25} (CVPR’25) for comparison with our ClusIR.

\begin{figure*}[!h]
    \centering
    \includegraphics[width=1\linewidth]{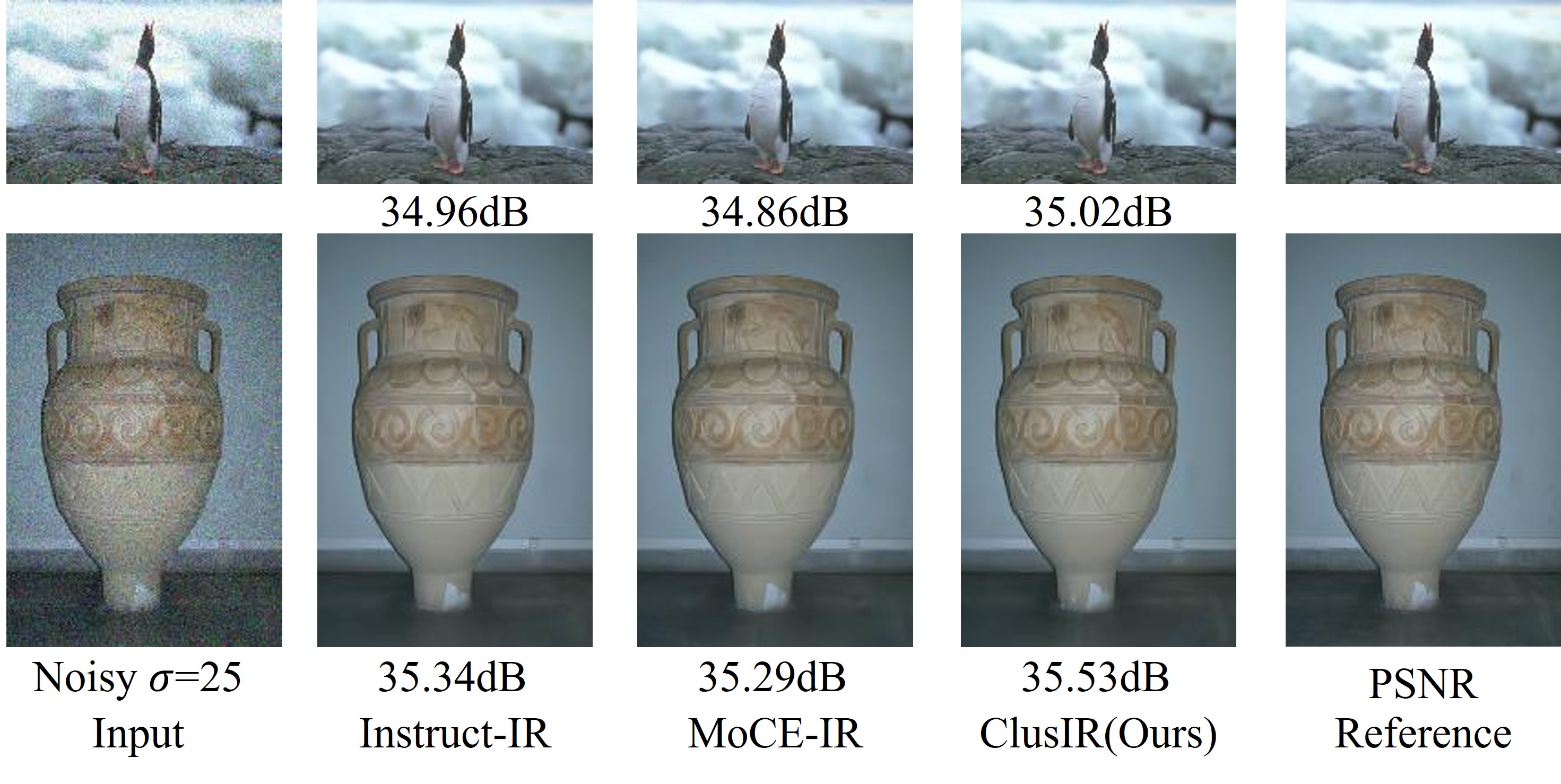}
    \vspace{-3em}
    \caption{Denoising  ($\sigma = 25$) visual comparisons of ClusIR with state-of-the-art All-in-One methods under ``\textcolor{blue}{\textbf{N}}+\textcolor{green}{\textbf{H}}+\textcolor{red}{\bf{R}}+\textcolor{purple}{\textbf{B}}+\textcolor{pink}{\textbf{L}}'' setting.}
    \label{fig:5tasks_denoise}
     \vspace{-1em}
\end{figure*}

\begin{figure*}[!h]
    \centering
    \includegraphics[width=1\linewidth]{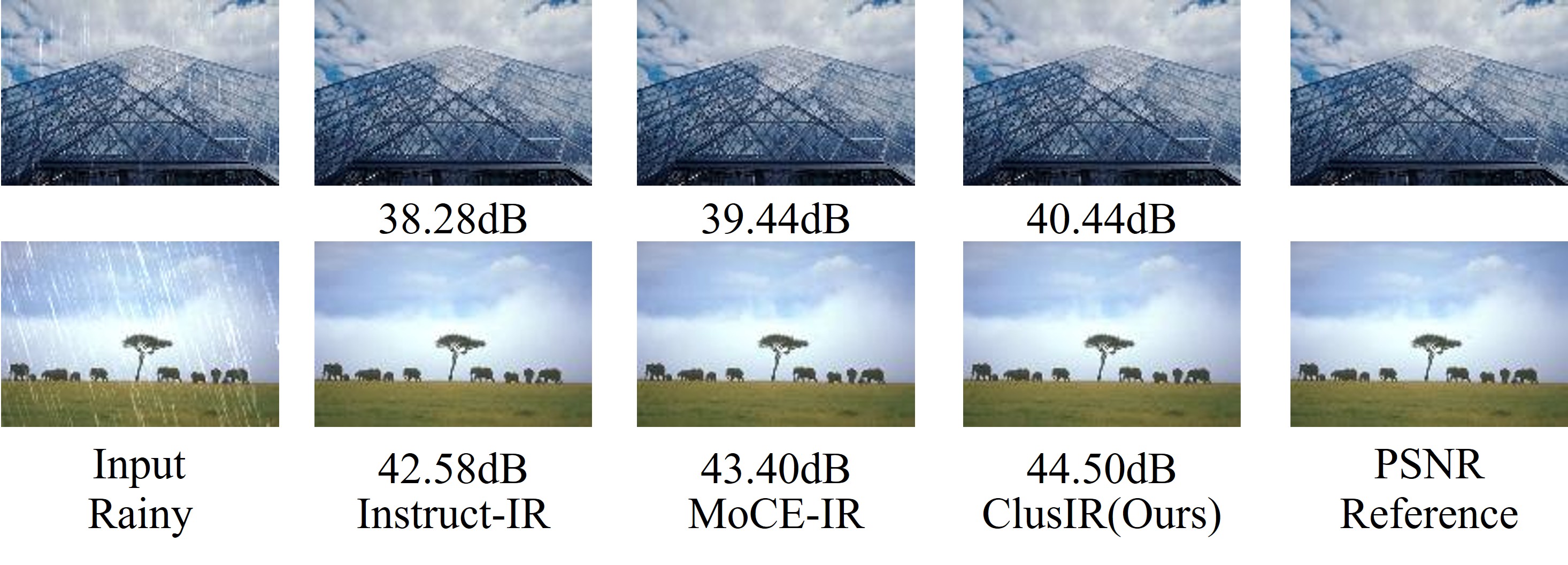}
   \vspace{-3em}
    \caption{Deraining visual comparisons of ClusIR with state-of-the-art All-in-One methods under ``\textcolor{blue}{\textbf{N}}+\textcolor{green}{\textbf{H}}+\textcolor{red}{\bf{R}}+\textcolor{purple}{\textbf{B}}+\textcolor{pink}{\textbf{L}}'' setting.}
    \label{fig:5tasks_derain}
    \vspace{-1em}
\end{figure*}

\begin{figure*}[!h]
    \centering
    \includegraphics[width=1\linewidth]{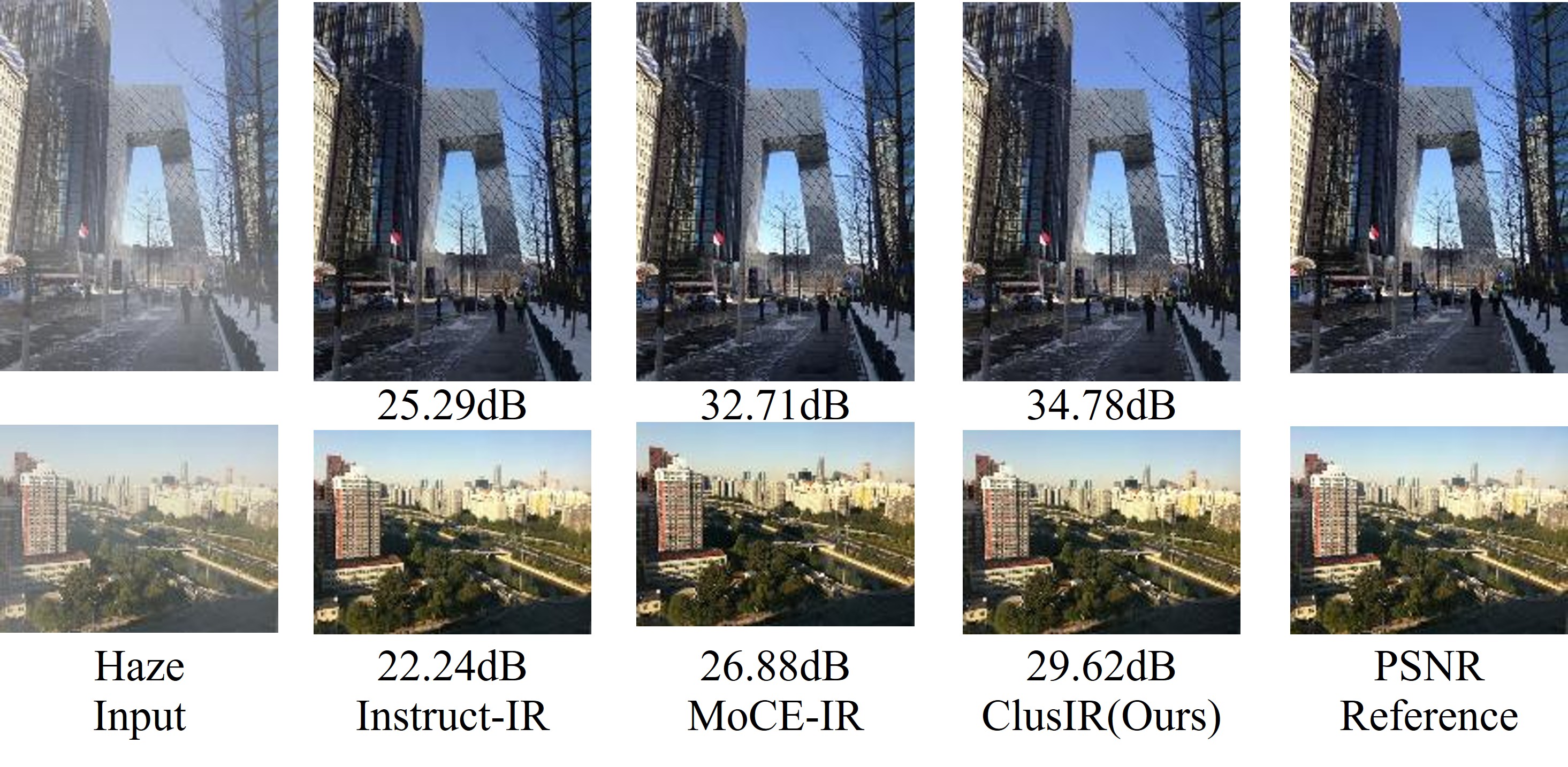}
    \vspace{-3em}
    \caption{Dehazing visual comparisons of ClusIR with state-of-the-art All-in-One methods under ``\textcolor{blue}{\textbf{N}}+\textcolor{green}{\textbf{H}}+\textcolor{red}{\bf{R}}+\textcolor{purple}{\textbf{B}}+\textcolor{pink}{\textbf{L}}'' setting.}
    \label{fig:5tasks_dehaze}
     \vspace{-1em}
\end{figure*}

\begin{figure*}[!h]
    \centering
    \includegraphics[width=1\linewidth]{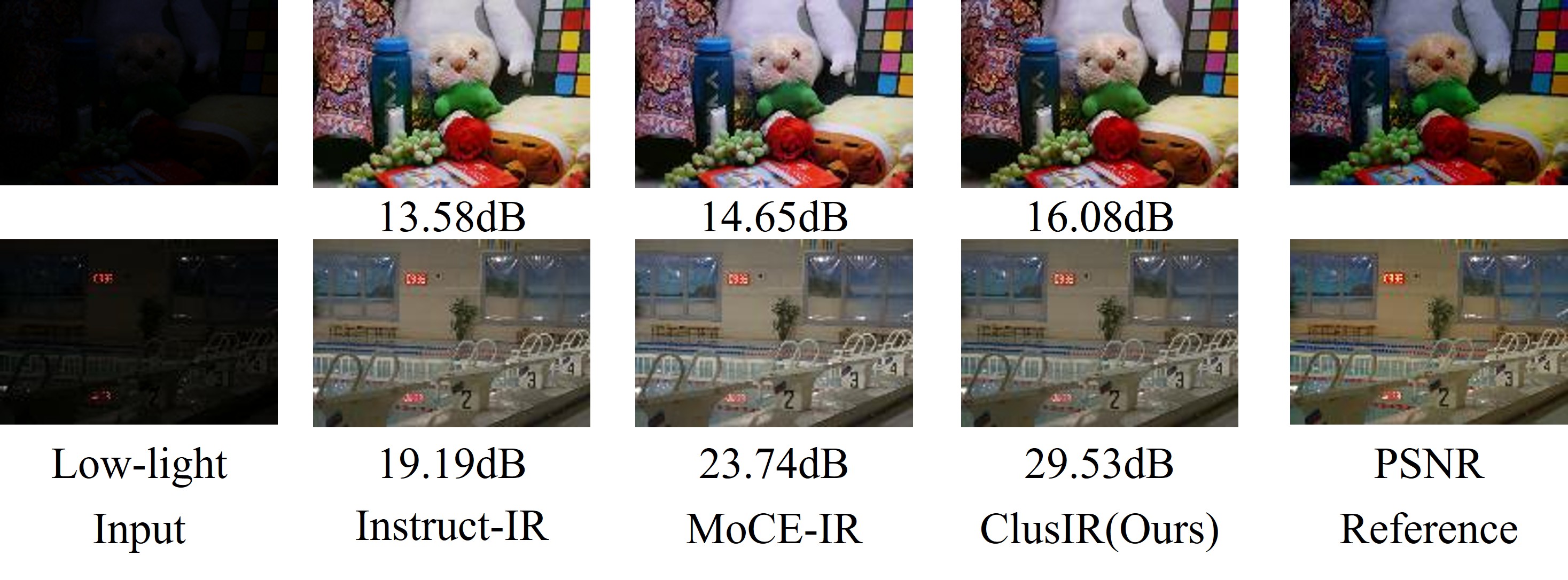}
    \vspace{-3em}
    \caption{Enhancement visual comparisons of ClusIR with state-of-the-art All-in-One methods under ``\textcolor{blue}{\textbf{N}}+\textcolor{green}{\textbf{H}}+\textcolor{red}{\bf{R}}+\textcolor{purple}{\textbf{B}}+\textcolor{pink}{\textbf{L}}'' setting.}
    \label{fig:5tasks_enhance}
     \vspace{-1em}
\end{figure*}

\begin{figure*}[!h]
    \centering
    \includegraphics[width=1\linewidth]{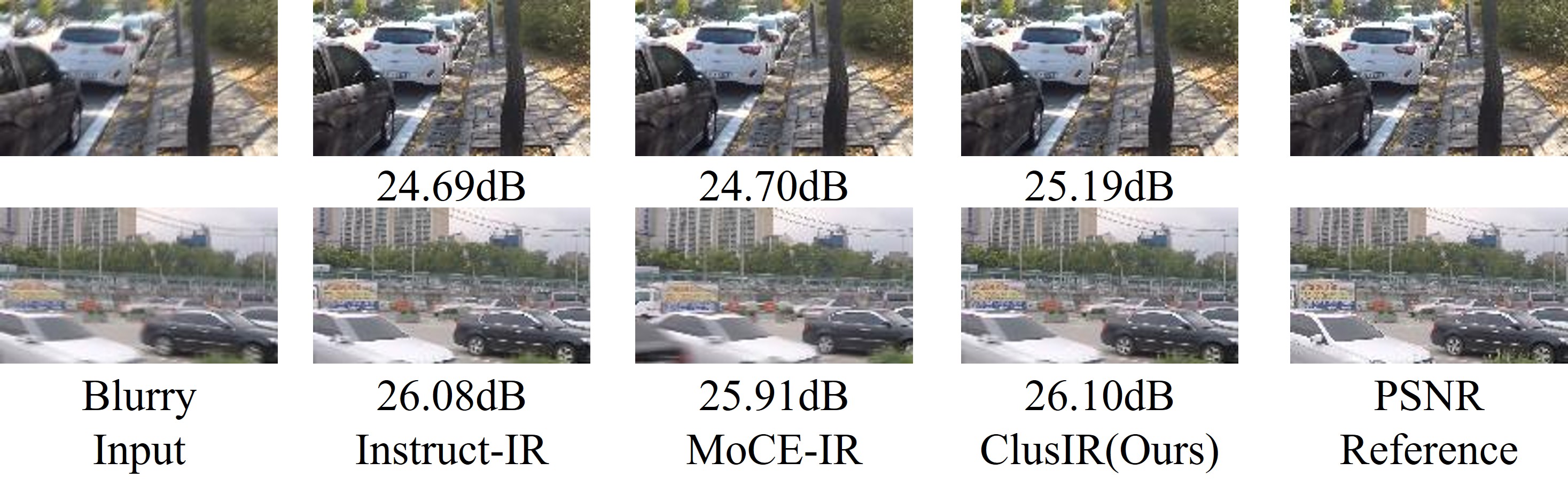}
    \vspace{-2.5em}
    \caption{Deblurring visual comparisons of ClusIR with state-of-the-art All-in-One methods under ``\textcolor{blue}{\textbf{N}}+\textcolor{green}{\textbf{H}}+\textcolor{red}{\bf{R}}+\textcolor{purple}{\textbf{B}}+\textcolor{pink}{\textbf{L}}'' setting.}
    \label{fig:5tasks_deblur}
\end{figure*}

Zoom-in views in Fig.~\ref{fig:5tasks_denoise}, \ref{fig:5tasks_derain}, \ref{fig:5tasks_dehaze}, \ref{fig:5tasks_enhance}, and \ref{fig:5tasks_deblur} reveal that ClusIR better preserves both textural and structural details, producing results that are visually closer to the reference images. These observations indicate better restoration quality compared to the competing methods. Further, we visualized the features after PCGRM-MoE under the complex scenario of the five-task degradation setting (Fig.~\ref{fig:5task_heatmap}). PCGRM achieved clear degradation feature discrimination in the shallow stages (i.e., stage 1-2), and gradually transitioned from degradation discrimination to semantic feature fusion in stages 3-4.
This demonstrates the effectiveness of the PCGRM in progressively refining and integrating degradation-specific features into semantic representations across multiple stages.

\subsection{Visual Comparison under Composited Degradations}
\begin{figure*}[!h]
    \centering
    \includegraphics[width=1\linewidth]{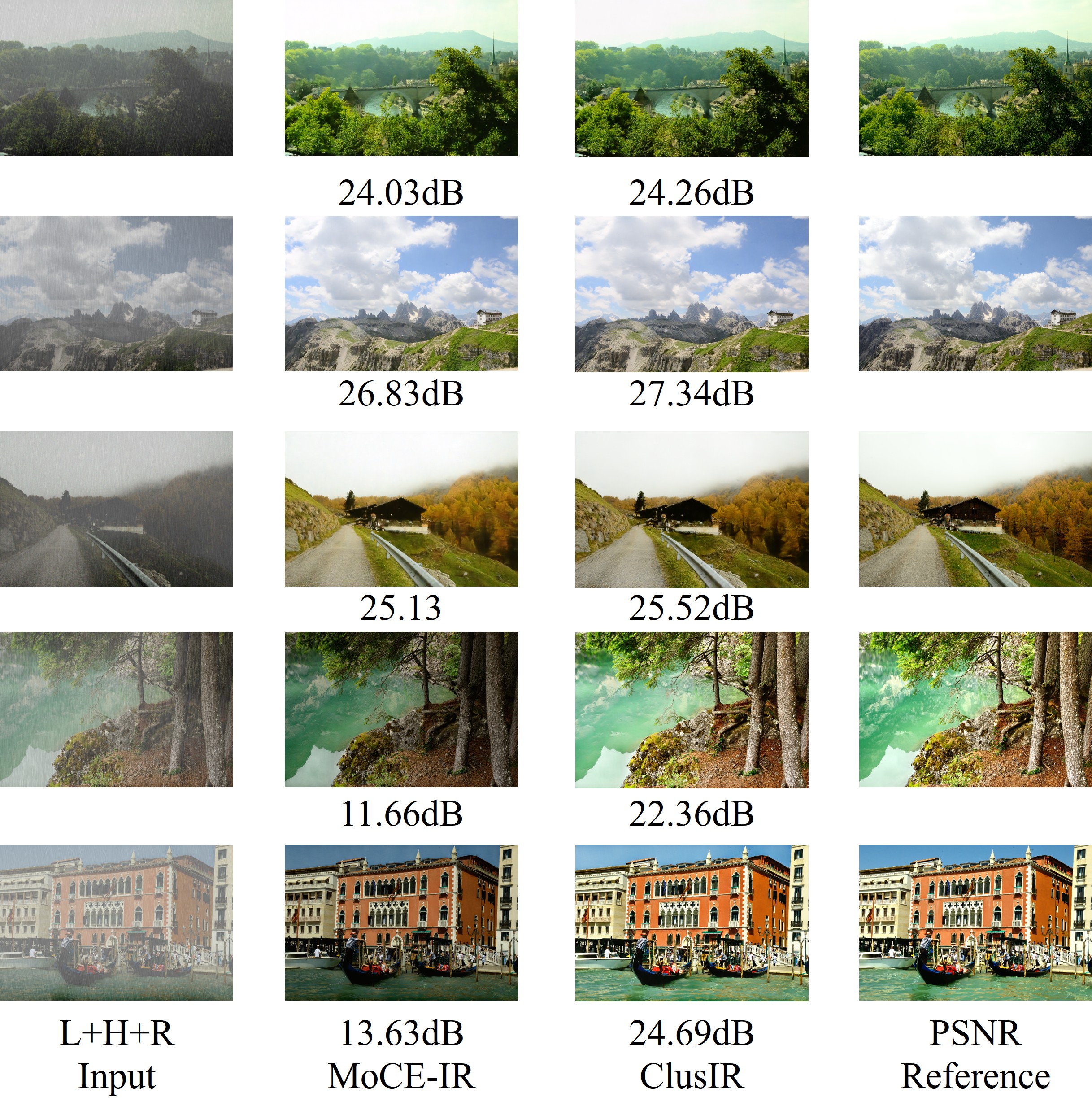}
    \vspace{-2.5em}
    \caption{Visual comparisons on the CDD11 dataset under the composite degradation setting.}
    \label{fig:com_lhr}
\end{figure*}

\begin{figure*}[!h]
    \centering
    \includegraphics[width=1\linewidth]{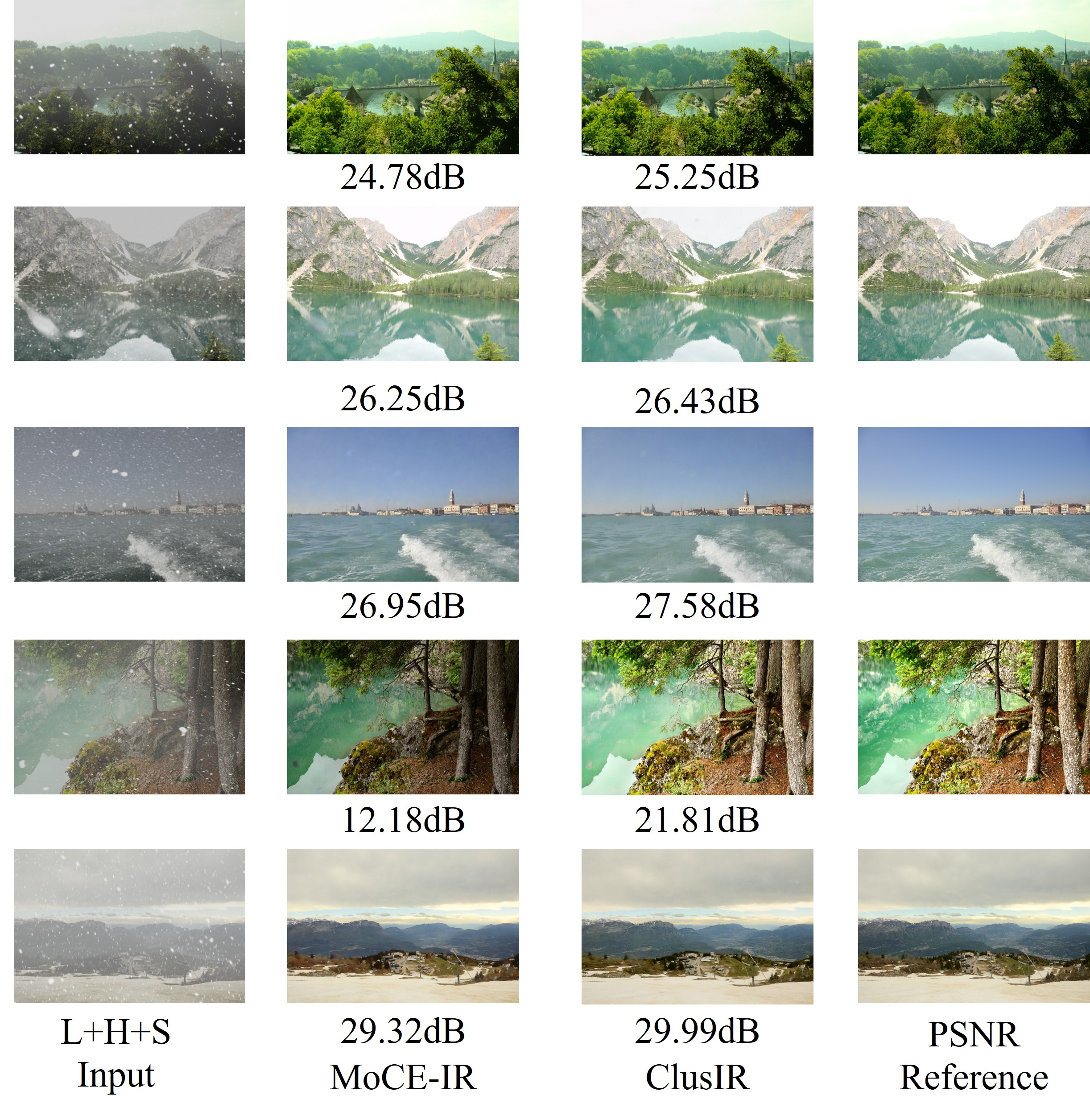}
    \vspace{-2.5em}
    \caption{Visual comparisons on the CDD11 dataset under the composite degradation setting.}
    \label{fig:com_lhs}
\end{figure*}

Fig.~\ref{fig:com_lhr} and Fig.~\ref{fig:com_lhs} present visual comparisons on the CDD11 dataset under composite degradation settings (L+H+R and L+H+S). 
Compared with MoCE-IR~\cite{moceir}, ClusIR consistently produces clearer and more visually consistent results, especially under challenging haze-dominated scenarios. 
These results demonstrate that under complex multi-degradation conditions, the proposed PCGRM enables ClusIR to disentangle degradation-specific representations, enabling more accurate restoration of clean image structures.

\subsection{Limitations}
Although ClusIR demonstrates strong robustness under coupled degradations, the proposed two-stage routing and frequency refinement design introduces additional computational overhead and inference latency compared to lightweight single-branch models. 
While this design effectively enhances degradation-aware feature modeling, it inevitably increases model complexity, making it less suitable for real-time or resource-constrained deployment scenarios.
 
\end{document}